\documentclass[11pt,letterpaper]{article}
\usepackage[T1]{fontenc}
\usepackage[utf8]{inputenc}
\usepackage{mathptmx}
\usepackage[scaled=0.92]{helvet}
\usepackage{courier}
\usepackage[margin=1in]{geometry}
\usepackage{microtype}
\usepackage{graphicx}
\usepackage{amsmath}
\usepackage{amssymb}
\usepackage{pifont}
\usepackage{longtable}
\usepackage{array}
\usepackage{algorithm}
\usepackage{algpseudocode}
\usepackage[dvipsnames]{xcolor}
\usepackage{fancyhdr}
\usepackage{caption}
\usepackage{float}
\usepackage{xurl}
\usepackage[colorlinks=true,linkcolor=RoyalPurple,citecolor=RoyalPurple,urlcolor=RoyalPurple]{hyperref}
\usepackage[capitalize,noabbrev]{cleveref}
\hypersetup{
  pdftitle={World-Grounded Reinforcement Learning for Pollution-Aware Bird Movement Modeling Using a Multi-Instance DQN Framework},
  pdfauthor={Shivin Mangal, Divyansh Chawla, Anshu Garg, Isshaan Singh},
  pdfkeywords={world-grounded reinforcement learning, deep Q-network, multi-instance simulation, air quality index, avian movement modeling, environmental monitoring}}

\newcommand{\venuenote}{Accepted at the GlobalSouthAI Affinity Event, NeurIPS 2026}
\newcommand{\venueurl}{https://neurips.cc/virtual/2026/loc/sydney/162655}
\fancypagestyle{firstpage}{\fancyhf{}\fancyfoot[C]{\thepage}}

\title{\vspace{-0.6em}\rule{\linewidth}{1.2pt}\\[0.6em]
  \textbf{World-Grounded Reinforcement Learning for Pollution-Aware Bird Movement Modeling Using a Multi-Instance DQN Framework}\\[0.3em]
  {\large Extended Version}\\[0.2em]
  \rule{\linewidth}{0.6pt}}
\author{%
  Shivin Mangal\textsuperscript{1}\quad
  Divyansh Chawla\textsuperscript{1}\quad
  Anshu Garg\textsuperscript{1}\quad
  Isshaan Singh\textsuperscript{2}\\[0.5em]
  \normalsize\textsuperscript{1}Vellore Institute of Technology (VIT), Chennai, India\qquad
  \textsuperscript{2}MBZUAI, Abu Dhabi, UAE
}
\date{}

\begin{document}
\maketitle
\thispagestyle{firstpage}

\begin{abstract}
Birds are widely used as bioindicators because their movement, physiology and habitat use respond to atmospheric pollution, which is summarised by the Air Quality Index (AQI). Conventional avian and environmental monitoring relies largely on static datasets, field observation and costly tracking, which limits the study of how movement could respond to spatially varying pollution. This paper presents a \emph{world-grounded} reinforcement learning (RL) framework in which measured environmental data, rather than a synthetic environment, supply the feedback signal for learning agents. India is discretised into a land-masked $109\times108$ grid of $0.27^\circ$ cells; a precomputed AQI matrix, assembled from sampled air-quality data and neighbour imputation, replaces repeated API queries during training and execution; and state and union-territory boundaries are rasterised into an encoded state map. A multi-instance Deep Q-Network (DQN) is trained over more than 400 agents initialised from 479 Xeno-canto bird-recording locations, using $\epsilon$-greedy exploration, experience replay, a target network and anti-stagnation heuristics. The agents navigate the AQI-integrated environment under a pollution-aware reward with boundary, out-of-scope and repetition penalties. We compare DQN with Monte Carlo, Q-Learning and Value Iteration on cumulative reward, action diversity, reward change rate, step efficiency and distance travelled, and analyse simulated traversal paths and learning error. In the reported runs DQN attains the highest cumulative reward and step efficiency, the per-step reward reaches a peak of 722, and the mean absolute error over the first ten agents averages 1.0716 (minimum 0.0291, maximum 5.1144). Because the recordings are occurrences rather than tracks and the AQI layer is static, the generated paths are simulated responses to a specified objective, not reconstructions of migration; we therefore present the framework as a computational proof of concept and decision-support basis, and outline the multi-seed, time-resolved and trajectory-based validation required for ecological conclusions. This extended version gives the complete formulation, algorithms, data description and analysis that were condensed in the workshop paper.
\end{abstract}

\begin{center}
\fcolorbox{RoyalPurple}{RoyalPurple!6}{%
\begin{minipage}{0.92\linewidth}
\centering\small
\textbf{A condensed version of this work has been accepted at the GlobalSouthAI Affinity Event at NeurIPS 2026.}\\[0.2em]
Workshop paper and poster: \url{\venueurl}\\[0.2em]
This arXiv preprint is the extended version, with the full methodology, algorithms, data analysis and discussion.
\end{minipage}}
\end{center}


\section{Introduction}

Understanding the effects of pollution and climate change requires moving beyond static observation toward learning frameworks that take measurable environmental consequences as feedback~\cite{rl_survey_2020}. Modern sensing and computational tools allow ecological systems to be analysed at scale, yet many approaches still rely on fixed models that do not capture the interaction between environmental conditions and behaviour~\cite{ecological_modeling_2014_2024}. Treating high-resolution pollution data as a feedback signal for adaptive learning systems offers a route to more responsive, consequence-aware modelling~\cite{rl_environmental_optimization_2026}. We refer to this setting as \emph{world-grounded} reinforcement learning: the agent's environment, and the reward it learns from, are built from measured geographic and air-quality data rather than from an abstract simulator.

The quantification and prediction of pollution and climate change impacts on ecosystems and natural hazards are essential for effective environmental monitoring and modeling. Advanced data collection and analytical techniques, utilizing highly sensitive sensors and computational models, are employed to estimate ecological processes, species distributions, and the frequency and intensity of natural hazards in response to environmental variations. Recent studies on pollution \cite{Chen2025,Tian2024} and climate change \cite{Mahdian2024,Ahmadi2024,Donnelly2024} emphasize the necessity of continuous monitoring and the compilation of holistic datasets, which facilitate trend identification and long-term ecosystem health assessments. Freshwater ecosystems play a crucial role in transporting land-based plastics into oceans, with microplastic particles exhibiting persistent and complex behaviors influenced by local hydrodynamics and water chemistry. Environmental monitoring and modeling techniques have been demonstrated to effectively capture these dynamics, including processes such as settling, aggregation, retention, and suspension \cite{Chen2025}. The accurate prediction of microplastic transport and fate underscores the need for comprehensive monitoring strategies to mitigate pollution. Additionally, studies on polyurethane plastic degradation in aquatic environments have highlighted the significant role of microorganisms and light exposure in the process \cite{Tian2024}. The interplay between microbial activity and photodegradation governs microplastic formation rates and their environmental impact, thereby providing critical insights for pollution management strategies.

In the context of climate change, research has quantified its effects on wetland shrinkage in Iran, predicting severe desiccation due to rising temperatures and declining precipitation levels \cite{Mahdian2024}. A novel gray-box modeling framework, integrating process-based models with machine learning techniques, has been employed to predict wetland shrinkage, illustrating the intricate relationships between climate variables and human activities. Such advanced modeling approaches are essential for forecasting and mitigating the impacts of climate change on critical ecosystems. Furthermore, research on resilient infrastructure design has assessed the effects of catastrophic floods on essential structures, such as bridges, using computational and predictive modeling tools \cite{Ahmadi2024}. The integration of computational fluid dynamics (CFD) with machine learning approaches has facilitated the simulation of hydrodynamic pressures and the evaluation of structural resilience against extreme flood events. The predictive accuracy of these models is significantly enhanced through the combined use of CFD and machine learning, offering valuable insights for designing and maintaining infrastructure amid climate-induced disasters. Advances in physics-informed machine learning have further led to the development of surrogate models for flood prediction, demonstrating robustness in data-limited contexts \cite{Donnelly2024}. By incorporating physical prior knowledge into neural networks, these models improve the efficiency and accuracy of flood simulations, supporting the development of real-time flood prediction models that enhance disaster preparedness and response.  

Environmental monitoring and modeling provide a comprehensive understanding of ecosystems and their responses to pollution and climate change. The integration of advanced computational techniques, machine learning, and traditional monitoring methods allows for a more dynamic and precise analysis of ecological changes \cite{Henheik2022}. Such methodologies facilitate accurate environmental predictions, enabling the formulation of effective mitigation strategies to enhance ecosystem resilience and sustainability in response to global environmental challenges.  

Bird species are recognized as critical bioindicators in environmental monitoring due to their sensitivity to ecological changes, making them reliable indicators of ecosystem health and biodiversity. The significance of tracking bird populations to assess the impacts of pollution and climate change has been underscored by Rigal et al.~\cite{Rigal2023}, who emphasize that fluctuations in bird populations reflect broader ecological shifts, providing valuable insights into ecosystem stability and environmental health. Insects, including beetles and butterflies, are also widely utilized as bioindicators to evaluate contamination levels across various environments \cite{Sanhita2023}, offering critical information regarding pollutant presence and concentration. Seabirds, in particular, serve as effective indicators of marine ecosystem health, given their heightened sensitivity to habitat changes \cite{Laini2022}. Physiological biomarkers in seabirds facilitate the assessment of environmental stressors and ecological pressures on populations \cite{Norte2022}, providing a means of evaluating exposure to chemical pollutants and other environmental disturbances. The diverse dietary and habitat requirements of bird species further reinforce their role as indicators of environmental change, illustrating the widespread impact of ecological stressors on ecosystems \cite{Milla2022}.  

Given the ecological significance of birds, simulating their movement under varying environmental conditions is a useful complement to field observation for anticipating ecological impacts. Modelling such scenarios on measured environmental data allows the exploration of how movement might respond to pollution and habitat availability, which can inform conservation strategies and management decisions. However, existing environmental monitoring systems predominantly rely on static datasets and lack the capacity for adaptive, interaction-driven modelling. Traditional observation and data collection techniques, although valuable, are resource-intensive and often fail to account for the complexities of ecological systems. Moreover, many current models, as discussed in subsequent sections, provide only a fragmented understanding of ecosystem interactions, lacking the integration of diverse data sources and advanced computational methodologies. Data-driven, learning-based modelling approaches are therefore a natural extension of environmental monitoring and conservation efforts.

By leveraging measured data and advanced computational techniques, simulations can offer a richer representation of ecosystem interactions. Such approaches can help identify candidate ecological thresholds and pressure points for closer study. Simulation models that integrate climate variables, pollutant levels, and habitat changes provide a structured way to assess their potential impact on bioindicator species, and can serve as decision-support tools for conservation strategies and evidence-based policy.

To address the limitations of conventional monitoring approaches, this work introduces a world-grounded simulation framework that uses multi-instance reinforcement learning to model bird movement across a large region. The framework integrates hybrid data types---latitude, longitude, altitude, and Air Quality Index (AQI) values---with AQI acting as the measured environmental signal \cite{Ware2023}. Deep Q-Network (DQN) agents learn to adjust their paths in response to AQI variations and geospatial constraints, making the agents' learning process and decision-making strategies observable and comparable across algorithms. By grounding the simulation in real geographic structure and measured air quality, the framework provides a data-driven basis for studying environmental interactions and for supporting conservation policy and management \cite{Xu2023}.

The simulations are grounded in real-world observations through bird data from the Xeno-canto database, which contains extensive bird call recordings along with associated metadata \cite{Xeno-canto}. The recording locations provide geographic context and initial states for the agents; they are occurrence records rather than movement tracks, so the trajectories produced by the framework are simulated responses to a specified pollution-aware objective. Environmental conditions are relevant to avian physiology and ecology~\cite{Norte2022}; here, a preference for cleaner cells is a modelling assumption rather than an empirically established movement rule, and AQI is used as a relative environmental-quality proxy that is not calibrated to species-specific exposure or fitness. The major contributions of this study are as follows:

\begin{itemize}
    \item \textbf{World-grounded AQI-integrated environment:} A country-scale environment for India that combines state and union-territory boundaries, bird-recording locations and a measured AQI layer on a land-masked $109\times108$ grid of $0.27^\circ$ cells. Habitat and environmental quality enter the agent's state and reward through latitude, longitude, AQI and an encoded state identifier.
    \item \textbf{Precomputed AQI matrix:} AQI values over the whole region are precomputed into a NumPy matrix using sub-cell sampling and neighbour imputation, replacing repeated API queries and making state transitions and reward computation fast enough for large-scale training.
    \item \textbf{Multi-instance DQN framework:} More than 400 agents, each initialised from a Xeno-canto record, are trained with a DQN using $\epsilon$-greedy exploration, experience replay, a target network and anti-stagnation heuristics (repetition penalties, weight restoration and exploration resets), together with environment functions for movement, reward computation and state updates.
    \item \textbf{Comparative evaluation and analysis:} DQN is compared with Monte Carlo, Q-Learning and Value Iteration on cumulative reward, action diversity, reward change rate, step efficiency and distance travelled, complemented by traversal-path, reward-trajectory and error analyses and by qualitative comparison with bioindicator, simulation and RL literature.
\end{itemize}

The remainder of the paper is organised as follows. Section~\ref{sec:background} reviews related work and formulates the problem. Section~\ref{sec:method} details the world-grounded environment, the AQI mapping, the DQN formulation, the multi-instance runtime and the data. Section~\ref{sec:results} presents the experimental setup, the evaluation metrics and the comparative results. Section~\ref{sec:related} compares the framework qualitatively with related work, and Section~\ref{sec:discussion} discusses implications and extensions. Section~\ref{sec:limitations} sets out limitations, validation requirements and future work, and Section~\ref{sec:conclusion} concludes.

\section{Background and Problem Formulation}\label{sec:background}

This section reviews the methodologies that underpin the proposed world-grounded simulation framework and the multi-instance reinforcement learning model for bird movement analysis. Relevant studies and technological advances are reviewed, and the problem is then formally stated, positioning the framework within the broader scientific landscape.

\subsection{Literature Review}

Effective environmental monitoring plays a crucial role in assessing ecosystem health and detecting changes driven by natural or anthropogenic factors. Traditional approaches, including satellite imagery and ground-level sensors, encounter limitations related to cost, spatial coverage, and real-time responsiveness \cite{Singh2023}. An alternative strategy involves utilizing bioindicator species, particularly birds, which offer dynamic insights into ecosystem health and pollution levels \cite{Larionova2023}. The integration of ecological monitoring with drone-enabled Internet of Remote Things (IoRT) has introduced a transformative approach, facilitating data collection in remote and extreme environments while enhancing the accuracy of environmental assessments \cite{Zhang2022}. Birds contribute significantly to ecological processes such as pollination, seed dispersal, and predation, as demonstrated in multiple studies \cite{EnvironmentalMonitoring2022,Platel2023,Ortega2023,Lapbaeva2023}. Their population dynamics and behavioral patterns serve as reliable indicators of habitat quality and environmental stressors, making them essential for the early detection of ecological disturbances \cite{Jonsson2023}. Research consistently establishes a strong correlation between avian health and environmental conditions, reinforcing their importance in conservation initiatives and biodiversity preservation.  

Bird movement and habitat selection are influenced by multiple environmental factors across species. Migration route preferences in passerines, such as the Golden-crowned Sparrow, are determined by vegetation types, climate conditions, and land cover characteristics \cite{Conete2022}. Urbanization alters avian fitness and behavior by modifying vegetation, temperature, and resource availability, subsequently affecting breeding success \cite{Iverson2023}. Free-ranging domestic dogs demonstrate habitat selection patterns that prioritize human-associated resources \cite{Chen2023}, while vultures select roosting sites based on urban cover and proximity to water, with diurnal roosts frequently located near human settlements \cite{Cunha2022}. Likewise, urbanization presents challenges for species such as the great tit, an insectivorous passerine, by diminishing food availability and increasing susceptibility to climatic variations, ultimately reducing reproductive success in urban environments \cite{Hill2023}. These findings collectively highlight the intricate relationship between bird movement, habitat selection, environmental factors, and human-induced landscape transformations. Understanding these dynamics is essential for developing effective conservation strategies and mitigating the adverse effects of environmental change.  

Agent-based models (ABMs) have been widely employed in ecological research to simulate interactions among individual agents within an ecosystem, facilitating the analysis of population dynamics and ecosystem processes \cite{Saulnier2023,Antelmi2022}. Recent advancements in deep learning, particularly deep reinforcement learning (DRL), have substantially enhanced ecological modeling by enabling the processing of large datasets and improving predictive accuracy, thereby providing deeper insights into ecosystem dynamics \cite{Ushio2023}. The introduction of computational frameworks such as ecological reservoir computing (ERC) has demonstrated that ecological networks possess computational capabilities, allowing real-time predictions and information processing based on population dynamics \cite{Perry2022}. The integration of ABMs with DRL has emerged as a promising approach for developing sophisticated simulation models capable of predicting complex system behaviors under varying conditions while enabling adaptive learning in response to environmental changes. This integration strengthens ecological studies and enhances conservation and management strategies.  

Deep reinforcement learning incorporates reinforcement learning principles with deep neural networks, allowing agents to optimize their behavior through trial-and-error interactions with the environment \cite{Haensel2023}. Applications of DRL in ecological research have successfully modeled the adaptive behaviors of species in response to environmental variations, such as bird movement influenced by fluctuating pollution levels \cite{Martinez2023}. Multi-agent reinforcement learning (MARL) systems have been employed to investigate the emergence of strategic behaviors in microscopic environments, emphasizing the role of temperature in shaping ecological responses \cite{Tovey2023}. Beyond ecological applications, DRL algorithms have been utilized in modern renewable energy systems to address uncertainty and complexity, optimizing control policies through Markov Decision Process (MDP) formulations \cite{Li2023}. Furthermore, automata learning techniques have been developed to model environmental behaviors under agent control, thereby improving the transparency and interpretability of decision-making processes in complex ecosystems \cite{Tappler2023}. These findings further highlight the broad applicability of DRL methodologies across various domains, demonstrating their potential to transform ecological modeling and environmental management.

\subsection{Problem Formulation}

Predicting bird trajectories based solely on location data and Air Quality Index (AQI) values presents significant challenges, particularly when utilizing supervised learning methods. A primary limitation arises from the difficulty of constructing a large, labeled dataset that accurately maps bird movements to AQI levels. Bird flight patterns—encompassing both migratory and localized behaviors—are influenced by a multitude of environmental, biological, and anthropogenic factors, including wind patterns, habitat fragmentation, food availability, breeding seasons, and climate fluctuations. These factors collectively contribute to bird movement in ways that cannot be fully determined using only AQI and location data. Supervised learning depends on predefined relationships within the training dataset, assuming that future bird movements will resemble past patterns. However, bird behavior exhibits high nonlinearity, often undergoing abrupt and unpredictable shifts due to factors such as sudden changes in air quality, wind conditions, or food availability. Capturing such unpredictability within a static training dataset remains challenging, leading to poor generalization and inaccurate trajectory predictions. Furthermore, generating an exhaustive dataset that meticulously labels each bird movement with corresponding AQI values is not only resource-intensive but also impractical. Many critical variables that shape avian behavior may remain unaccounted for, making it difficult to create a sufficiently comprehensive training set. Consequently, a purely supervised learning approach is likely to oversimplify the problem, resulting in error-prone models with limited predictive accuracy for bird trajectories.  

Unsupervised learning, particularly clustering techniques such as K-means, has been explored to uncover hidden patterns in bird movement data without requiring labeled datasets. By analyzing AQI and geographic data, clustering enables the grouping of birds based on movement similarities, potentially revealing trends and correlations between environmental conditions and avian trajectories. However, significant limitations arise in this context. A primary drawback of unsupervised clustering methods, including K-means, is their static nature. These techniques assign bird movements to predefined clusters based on existing features such as AQI levels, latitude, and longitude. While this approach highlights general movement patterns, it fails to capture the temporal and dynamic nature of bird behavior. Birds respond to environmental changes in real time, a factor that clustering methods are not designed to accommodate. Additionally, clustering remains inherently retrospective—it classifies past behavior but lacks predictive capabilities for future movements. This rigidity represents a major limitation in bird movement prediction, as it does not account for the continuous, adaptive decision-making processes that birds employ in response to evolving environmental conditions. Another limitation of K-means and similar clustering approaches is their dependence on a fixed number of movement patterns or behavioral clusters, which oversimplifies avian behavior. Bird movement is inherently fluid, shaped by multiple fluctuating variables that cannot be easily categorized into a finite set of clusters. Furthermore, clustering algorithms typically assign equal weight to all features, failing to consider that certain environmental factors—such as sudden pollution spikes or abrupt weather changes—may exert a more immediate and significant impact on bird movement than others. Although unsupervised learning provides exploratory insights, it lacks the adaptability and predictive capabilities required for modeling bird movements in a dynamic environment.  

As summarized in Table~\ref{tab:advantages_disadvantages}, various indirect approaches have been employed for environmental health monitoring through external indicators. Biomarkers have been utilized to assess population dynamics and measure the impact of targeted interventions in both humans and flowers \cite{Larionova2023,Jonsson2023}. The integration of IoT-based systems has further facilitated remote environmental health assessments by leveraging technology for real-time ecosystem monitoring. Additionally, hybrid approaches that combine technological tools with natural indicators have demonstrated improved monitoring capabilities \cite{Iverson2023,Cunha2022,Hill2023}. Computational models are often employed to simulate ecological and evolutionary processes; however, their effectiveness is constrained by data availability and computational resource limitations \cite{Ushio2023}. Agent-based models (ABMs) are preferred in scenarios where well-defined decision-making processes exist, allowing for detailed simulations of individual and population-level interactions. Recent advancements have enabled the direct application of Deep Reinforcement Learning (DRL) to ecological and environmental tasks, yielding significant performance improvements \cite{Martinez2023,Tappler2023}. Some studies have also developed phenomenological models to replicate natural processes \cite{Malinowska2023}, while integrated frameworks that consider both static and dynamic environmental parameters have been employed to analyze bird activity in relation to multiple variables and movement patterns \cite{Ellis2023,Gong2025}. Building on these insights, this study introduces a novel approach to environmental monitoring, utilizing birds as indirect yet highly informative indicators of ecological conditions. Unlike traditional methods that involve physical tagging or sensor-based tracking, this approach computationally models bird behavior through simulated agents, incorporating AQI data and observational records. Reinforcement learning (RL) models should not replace conventional ecological monitoring techniques but rather complement them to enhance predictive accuracy and scalability. Established methods, including field observations, GPS tracking, and remote sensing, provide essential empirical data necessary for validating AI-driven models. However, these approaches are often limited by resource constraints, observational biases, and difficulties in capturing dynamic ecological interactions across large spatial and temporal scales. By integrating RL models with traditional techniques, avian movement patterns can be simulated, responses to varying AQI conditions can be predicted, and long-term environmental impacts can be analyzed with greater precision. This hybrid approach ensures that real-world ecological complexities are incorporated into computational frameworks while leveraging AI’s capability to efficiently process vast datasets. Consequently, RL-based systems should function as decision-support tools, augmenting traditional monitoring rather than replacing well-established ecological methodologies.  

This modelling framework is based on phenomenological principles and structured assumptions regarding the influence of pollution on movement. By grounding bird movement simulations in measured AQI data and real geography, it offers a scalable computational basis for environmental monitoring studies and a foundation on which ornithological factors can later be incorporated.

After evaluating the limitations of both supervised and unsupervised learning approaches, their inadequacy for modelling movement as a sequential decision process in a spatially varying environment becomes evident. Given the complexity of avian behaviour, which is shaped by many environmental and ecological factors, an approach is required in which behaviour emerges from interaction with the environment rather than from a fixed labelled mapping. We therefore formulate movement as a sequential decision problem in a world-grounded environment and solve it with Deep Reinforcement Learning. Deep RL enables an agent to learn movement policies through repeated interaction with the environment, making decisions from environmental feedback such as variations in AQI. The subsequent sections describe how the proposed DQN-based framework is constructed and evaluated, and how it complements traditional methods for studying bird propagation.

\begin{table}[p]
\centering
\small
\begin{tabular}{|c|p{3.4cm}|p{5.0cm}|p{5.0cm}|}
\hline
\textbf{Ref.} & \textbf{Method} & \textbf{Advantages} & \textbf{Disadvantages} \\
\hline
\cite{Larionova2023} & Biomonitoring for Metal Pollution & Evaluates metal effects via biological specimens. Identifies high-risk areas. & Limited by sample availability. Costly atomic absorption spectrometry. \\
\hline
\cite{Zhang2022} & Drone-Enabled IoRT & Real-time monitoring in remote areas. Reduces human risk. & High cost. Impacted by weather and regulations. \\
\hline
\cite{Jonsson2023} & eDNA for Bird Pollination & Non-invasive. Works across ecosystems. & DNA degradation affects accuracy. Needs extensive databases. \\
\hline
\cite{Iverson2023} & GPS-Based Bird Tracking & High-resolution migration insights. Supports conservation. & Expensive. Data gaps due to environmental factors. \\
\hline
\cite{Cunha2022} & GPS Tracking of Free-Roaming Dogs & Reveals habitat reliance. Aids rabies control. & Assumes data represents all dogs. Region-specific. \\
\hline
\cite{Hill2023} & Vulture Roost-Site Analysis & Identifies nocturnal vs. diurnal patterns. Helps urban planning. & Site-specific findings. May not reflect natural behaviors. \\
\hline
\cite{Ushio2023} & Ecological Reservoir Computing & Predicts nonlinear ecological trends. Integrates real-world data. & Lags in real-time applications. Needs extensive validation. \\
\hline
\cite{Haensel2023} & Agent-Based Learning Models & Enhances learning via simulations. Promotes critical thinking. & Complexity can overwhelm beginners. \\
\hline
\cite{Martinez2023} & DRL for Flocking & Minimal parameter tuning. Enhances coordination. & High computational cost. Long training times. \\
\hline
\cite{Tappler2023} & CASTLE for RL & Transparent decision-making. Adapts to continuous dynamics. & Slow without GPU. Suboptimal reward optimization. \\
\hline
\cite{Malinowska2023} & Virtual Species Modeling & Enhances species distribution models. Aids conservation. & Needs high-quality spatiotemporal data. \\
\hline
\cite{Emmet2023} & Bird Occupancy Modeling & Reduces spatial bias. Merges multiple datasets. & Hard to harmonize different surveys. \\
\hline
\cite{Yang2023} & CTMM-based Analysis & Reveals hidden interactions. Aids disease and behavior studies. & Model errors can affect accuracy. \\
\hline
\cite{Ellis2023} & Species Distribution Prediction & Captures dynamic habitat use. & Limited transferability. Needs extensive validation. \\
\hline
\cite{Gong2025} & Avian Radar Forecasting & Uses weather \& radar data for prediction. Machine learning boosts accuracy. & Hard to reconcile nocturnal data with weather. Needs simplification. \\
\hline
\end{tabular}
\caption{\label{tab:advantages_disadvantages}Indirect environmental-monitoring and modelling approaches reviewed in this work, with their main advantages and disadvantages.}
\end{table}

\section{Methodology: A World-Grounded Multi-Instance DQN Framework}\label{sec:method}

\begin{figure}[htbp]
    \centering
    \includegraphics[width=0.75\textwidth]{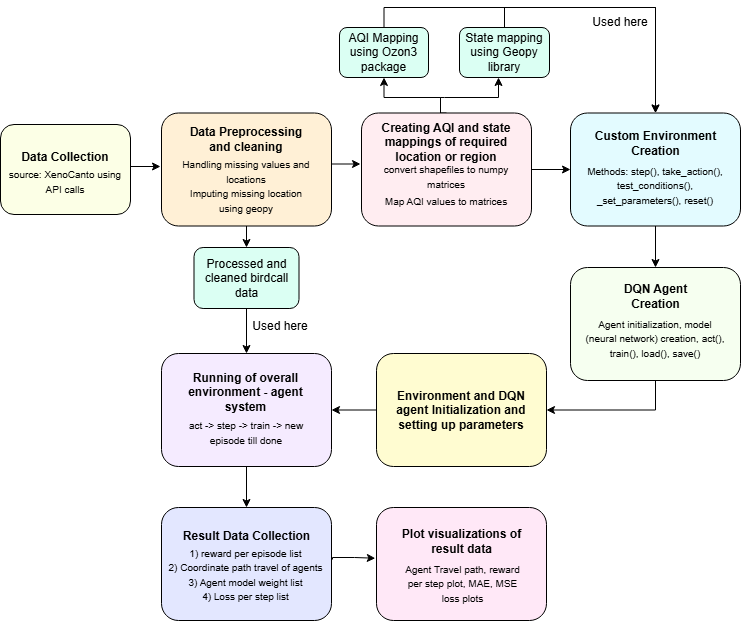}
    \caption{Overview of the world-grounded multi-instance DQN framework: data collection, spatial discretisation and masking, the precomputed AQI matrix, the AQI-integrated environment, DQN training and evaluation.}
    \label{fig:j1}
\end{figure}

In the proposed framework, bird-recording data are retrieved from the Xeno-canto database via API calls, followed by preprocessing that handles missing values and imputes state--city locations to ensure accurate geographic referencing. Bird locations are georeferenced using a shapefile of the study region, and a state-level map is obtained by rasterising the shapefile into a NumPy array, giving a structured representation of state boundaries.
To ground the environment in measured air quality, a precomputed AQI matrix is generated on a grid. Specifically, a \(109 \times 108\) grid is constructed from the country's coordinate bounds, and AQI values for each grid cell are obtained with the Ozon3 library~\cite{ozon3_github}. The resulting NumPy matrix supports state transitions and reward computation in the reinforcement learning framework without frequent API queries. A state-level mask restricts the environment to the region of interest, excluding neighbouring countries, oceans and other non-relevant areas. These spatial layers together form the world-grounded RL environment.
The framework trains many agents with a Deep Q-Network (DQN) in a multi-instance setting. Each agent is initialised from an individual bird record in the Xeno-canto dataset and navigates a structured environment in which the AQI map and state-level geography shape movement. Agents first explore the environment to learn its spatial structure and then exploit what they have learned. The environment is initialised with bird-recording metadata, AQI values and state boundaries.
Agents operate on a 2D grid map and can move in four directions---north, south, east and west. Each action triggers a state transition governed by latitude, longitude and AQI. The environment is defined by the functions \texttt{step}, \texttt{take\_action}, \texttt{test\_conditions} and \texttt{reset}, which together regulate movement, reward assignment and state updates. The reward reinforces favourable actions and penalises undesirable movements, guiding agents toward behaviour that satisfies the pollution-aware objective. An $\epsilon$-greedy policy balances exploration and exploitation, and, for learning stability, the DQN uses replay memory and two networks, with the target network periodically synchronised to the primary network. Figure~\ref{fig:j1} gives an overview of the methodology.

\subsection{World-Grounded AQI Mapping}

The Air Quality Index (AQI) summarises key atmospheric pollutants, including particulate matter (PM$_{2.5}$ and PM$_{10}$), ozone (O$_3$), sulfur dioxide (SO$_2$), nitrogen oxides (NO$_x$), and carbon monoxide (CO), which are known to affect avian physiology and behaviour. Fine particulate matter can penetrate deep into birds' respiratory systems, leading to oxidative stress and respiratory complications. Elevated ozone levels can disrupt oxygen exchange, impairing flight endurance and migratory efficiency. Sulfur dioxide and nitrogen oxides contribute to acid deposition, which affects food availability and nesting habitats, while carbon monoxide exposure interferes with oxygen transport. These pollutants can collectively influence migratory routes, breeding success and habitat selection, which motivates AQI as the environmental signal of the framework. In this work AQI is used as a relative measure of environmental quality that feeds the reward; it is not calibrated to species-specific exposure or fitness.

The methodology combines AQI mapping and state mapping so that the RL environment can be constructed for any geographic location for which boundary and air-quality data are available. The environment represents geographic regions through their environmental conditions, focusing on AQI and geographic parameters.

In the RL environment, states are defined by longitude, latitude and AQI, and AQI changes as the agent moves across the region. AQI values can be retrieved with the Ozon3 library~\cite{ozon3_github}, which provides atmospheric and pollution data for fixed locations. However, querying the API at every step is computationally expensive and subject to delays and rate limits. To avoid this and improve runtime efficiency, a precomputed AQI mapping of the study region is used.

The process of creating this mapping involves two key tasks: grid generation and AQI updating. A grid is defined over the geographical boundaries of the region, converting geographical locations (latitude and longitude) into discrete grid cells. The mathematical transformation from geographical coordinates to grid cells is represented as:

\begin{equation}\label{eq:gridx}
    x = \frac{\text{longitude} - \text{min}_{\text{long}}}{\Delta \text{long}}
\end{equation}

\begin{equation}\label{eq:gridy}
    y = \frac{\text{latitude} - \text{min}_{\text{lat}}}{\Delta \text{lat}}
\end{equation}

where $\text{min}_{\text{long}}$ and $\text{min}_{\text{lat}}$ are the minimum longitude and latitude in the shapefile of the selected region, and $\Delta \text{long}$ and $\Delta \text{lat}$ are the grid-cell sizes, approximately 0.27$^\circ$. The indices $x$ and $y$ identify the grid cell of a geographic coordinate. Equations~\eqref{eq:gridx} and~\eqref{eq:gridy} define the cells of a 2D matrix covering the region of interest.

Each grid cell is assigned specific longitude and latitude coordinates, later used to assign AQI values. The grid has a $109 \times 108$ structure, with each cell representing a 0.27-degree change in longitude and latitude, approximately 30 square kilometers. This grid is then masked to include only areas within the targeted region, excluding neighboring countries, oceans, and seas.

After establishing the grid, the AQI for each cell is calculated. Instead of querying AQI data for each coordinate in real-time during the simulation, a pre-computed AQI mapping matrix is used. The center points of the grid cells are queried to determine the average AQI for each cell:

\begin{equation}\label{eq:aqicell}
    \text{AQI}_{\text{cell}} = \frac{1}{n} \sum_{i=1}^{n} \text{AQI}_i
\end{equation}

where $\text{AQI}_{\text{cell}}$ is the AQI assigned to a grid cell, $\text{AQI}_i$ is the AQI of each sampled sub-cell or point within the cell, and $n$ is the number of sub-cells sampled. If data for a cell are unavailable, the missing value is approximated by averaging the values of the neighbouring cells:

\begin{equation}\label{eq:aqiapprox}
    \text{AQI}_{\text{approx}} = \frac{\text{AQI}_{\text{left}} + \text{AQI}_{\text{right}} + \text{AQI}_{\text{up}} + \text{AQI}_{\text{down}}}{4}
\end{equation}

Equation~\eqref{eq:aqiapprox} maintains spatial continuity of the AQI map in the presence of missing data points. The AQI values are retrieved iteratively with the Ozon3 library~\cite{ozon3_github}, with delays and retry attempts to respect rate limits. The resulting matrix is static: it represents spatial heterogeneity in air quality at the time of sampling, not temporal pollution dynamics.

For state mapping, the shapefile is rasterised into a binary matrix, where 1 marks regions of interest and 0 marks areas outside the study scope. Each state or union territory is uniquely encoded as:

\begin{equation}
    \text{State\_Code}(S) = i, \quad \text{where } i = 1,2,3,\dots,n
\end{equation}

where $S$ is a region of interest and $i$ is its unique encoded value. In the RL environment, each agent's state is represented as:

\begin{equation}
    \text{State}_{\text{agent}} = (\text{lat}, \text{long}, \text{AQI}, \text{State\_Code})
\end{equation}

Transitions between these states are made by the agent according to its learned policy. The base pollution-aware reward for transitioning into a grid cell is:

\begin{equation}\label{eq:rneg}
    R(s,a) = -\text{AQI}(s)
\end{equation}

where $R(s,a)$ is the reward for transitioning into state $s$ after taking action $a$, and $\text{AQI}(s)$ is the AQI of that state. The negative sign penalises entering states with higher AQI, encouraging the agent to avoid areas with poor air quality.

\begin{algorithm}[htbp]
    \caption{Precomputed AQI mapping for the world-grounded RL environment}
    \label{alg:aqi_mapping}
    \begin{algorithmic}[1]
        \State \textbf{Input:} Map grid with longitude and latitude coordinates
        \State \textbf{Output:} AQI mapping grid for RL framework
        
        \Function{CreateAQIMappingGrid}{}
        \State Divide the input map into a $109 \times 108$ grid.
        \State Calculate latitude and longitude change per cell ($\approx 0.27$ degrees).
        \State Determine area covered per cell ($\approx 30$ sq. km).
        
        \For{each cell in the grid}
            \State Divide the cell into $10 \times 10$ sub-cells.
            \State Compute latitude and longitude change per sub-cell ($\approx 0.09$ degrees).
            \State Determine area per sub-cell ($\approx 10$ sq. km).
            \For{each sub-cell}
                \State Compute AQI using mid-point latitude and longitude.
            \EndFor
            \State Compute the average AQI for all sub-cells within the cell.
        \EndFor
        
        \For{each cell}
            \State Count sub-cells inside and outside the region of interest.
            \If{sub-cells inside region $>$ sub-cells outside}
                \State Assign AQI as computed average.
            \Else
                \State Set AQI as -1 (insufficient data).
            \EndIf
        \EndFor
        
        \State \Return AQI mapping grid
        \EndFunction
    \end{algorithmic}
\end{algorithm}

Using Algorithm~\ref{alg:aqi_mapping}, the AQI mapping is generated and stored in a NumPy matrix in which each entry is the AQI of one grid cell. Each cell corresponds to a defined geographic region, giving a granular representation of air quality. This matrix is the environmental layer of the world-grounded RL framework, linking the agents' decisions to measured environmental conditions.

\subsection{AQI-Integrated Environment}

The AQI mapping is integrated with a multi-instance reinforcement learning framework trained with Deep Q-Learning (DQN). Agents are initialised from bird records in the Xeno-canto dataset and navigate an environment built from geographic and AQI data, so that their decisions depend on both location and air quality.

Birds are widely regarded as bioindicators of air quality because of their sensitivity to atmospheric pollutants, rapid physiological responses and extensive geographic distribution. Their highly efficient yet vulnerable respiratory systems mean that airborne contaminants such as particulate matter (PM\(_{2.5}\), PM\(_{10}\)), ozone (O\(_3\)), sulfur dioxide (SO\(_2\)) and nitrogen oxides (NO\(_x\)) can affect them more rapidly than mammals. Changes in bird populations, migratory behaviour and physiology can therefore provide early indications of deteriorating environmental conditions, and, because birds occupy various trophic levels and habitats, their responses can reflect broader ecological health. This motivates bird-recording locations as the anchor points of the simulation.

During the exploration phase, agents traverse the environment and gather information about different states, defined as combinations of geographic coordinates and AQI. After sufficient exploration, agents exploit the acquired knowledge to move toward states that yield higher cumulative reward. The environment is constructed from three datasets:
\begin{itemize}
    \item Bird-recording metadata from Xeno-canto
    \item The precomputed AQI matrix
    \item State and union-territory boundary data
\end{itemize}
Together these datasets ground the environment in real geographic and environmental variation. The AQI matrix provides measured air-quality information, while the state data define geographic boundaries as polygons in latitude and longitude.

The action space for agents is defined as:
\begin{equation}
    \text{Action Space} = \{\text{North, South, East, West}\}
\end{equation}

The observation space consists of three features: latitude, longitude and AQI. The environment defines the step, take-action, test-conditions and set-parameters functions outlined in Algorithm~\ref{algo:bird_propagation}.

The reward at time step $t$ is computed as:
\begin{equation}\label{eq:raff}
    R_t = \beta - \alpha \times \text{AQI}_t
\end{equation}
where $\alpha$ is the penalty weight for poor air quality and $\beta$ is a constant base reward for maintaining favourable AQI levels.

Agent movement updates are defined as:
\begin{align}
    \text{New Latitude} &= \text{Current Latitude} \pm 0.27 \\
    \text{New Longitude} &= \text{Current Longitude} \pm 0.27
\end{align}

AQI acceptance bounds are evaluated relative to the agent's reference AQI:
\begin{align}
    \text{Lower Limit} &= 0.25 \times \text{AQI}_{\text{agent}} \\
    \text{Upper Limit} &= 1.40 \times \text{AQI}_{\text{agent}}
\end{align}

and the AQI-based reward is:
\begin{equation}\label{eq:rband}
    R_{\text{AQI}} =
    \begin{cases}
        +7, & \text{if } \text{Lower Limit} \leq \text{AQI}_{\text{Location}} \leq \text{Upper Limit} \\
        -5, & \text{otherwise}
    \end{cases}
\end{equation}

\paragraph{Relationship between the reward expressions.} The affine reward~\eqref{eq:raff} reduces to the negative-AQI reward~\eqref{eq:rneg} when $\alpha=1$ and $\beta=0$, provided both evaluate AQI at the same point of the transition. The acceptance-band rule~\eqref{eq:rband} is qualitatively different: it rewards remaining within a band relative to the agent's reference AQI, so it can reward an AQI increase of up to 40\% and penalise a reduction below 25\% of the reference. It therefore expresses a preference for conditions similar to the agent's reference rather than a monotonic preference for cleaner cells. These thresholds are simulation design choices, not validated avian tolerance limits, and reward magnitude alone does not establish reduced pollution exposure.

\begin{algorithm}[htbp]
    \caption{World-grounded bird propagation environment (AQI-integrated, DQN approach)}
    \label{algo:bird_propagation}
    \begin{algorithmic}[1]
        \State \textbf{Input:} 
        \State \quad Data: Bird data (latitude, longitude, altitude, AQI)
        \State \quad Country: Geopandas DataFrame containing region shapefile
        \State \quad State Data: Dictionary of states and boundary coordinates
        \State \quad AQI Matrix: AQI mapped data of the region
        \State \textbf{Output:}
        \State \quad (a) Total reward accumulated
        \State \quad (b) Traversal coordinates for each agent
        \State \quad (c) New state values after taking action
        \State \quad (d) Done flag for execution status
        \State \quad (e) Finish flag for completion status
        
        \Function{InitializeEnvironment}{}
            \State Initialize the AQI-integrated environment with data, state data, and AQI matrix
            \State Define action space: Discrete(4) \{North, South, East, West\}
            \State Set observation space with latitude, longitude, and AQI features
            \State Initialize parameters with \textit{reset()}
            \State Set finish flag to \textit{False}
        \EndFunction
        
        \Function{StepFunction}{action}
            \If{episode is complete}
                \If{all rows processed}
                    \State Reset index and return results
                \Else
                    \State Increment step index, reset parameters, and update state
                \EndIf
            \EndIf
            \State Execute action by calling \textit{take\_action(action)}
            \State Compute reward using \textit{test\_conditions()}
            \State \Return new state, traversal coordinates, reward, and completion status
        \EndFunction
        
        \Function{TakeAction}{action}
            \State Update latitude or longitude based on action
            \State Append new coordinates to travel list
            \State Update state with new latitude, longitude, and altitude
        \EndFunction
        
        \Function{TestConditions}{}
            \State Retrieve AQI value for current step
            \If{latitude/longitude out of bounds}
                \State Reset parameters and return penalty
            \EndIf
            \If{AQI location is zero (out of scope)}
                \State Return penalty
            \EndIf
            \State Compute AQI range and check if within limits
            \State Assign rewards accordingly
        \EndFunction
        
        \Function{SetParameters}{}
            \State Initialize agent latitude, longitude, and altitude
            \State Track traversal paths and cumulative rewards
        \EndFunction
        
        \Function{ResetEnvironment}{}
            \State Reset state list, longitude, latitude, altitude
            \State Reset location, done flag, and travel list
        \EndFunction
    \end{algorithmic}
\end{algorithm}

The environment handles a discrete action space with a structured state representation. Several challenges nonetheless arise, including computational demand, the size of the state space and the complexity of reward design. Q-learning provides a way to learn policies without an explicit model of the environment, and Deep Q-Learning (DQN) extends it by approximating Q-values with a deep neural network, which allows the agents to generalise across the continuous-valued state of latitude, longitude and AQI.

\subsection{DQN Model Formulation}

The DQN approach is used to model bird propagation over the AQI map. The agent learns to navigate an environment defined by coordinates and AQI values, with the objective of maximising cumulative reward by reaching and remaining in regions that satisfy the pollution-aware objective. The environment is a 2D grid on which agents move in one of four cardinal directions, as detailed in the previous subsection. At each time step $t$, the agent's state $s_t$ is:

\begin{equation}
    s_t = [\text{latitude}, \text{longitude}, AQI]
\end{equation}

The agent selects an action using an epsilon-greedy policy, balancing exploration and exploitation. The probability of selecting a random action decreases over time, following:

\begin{equation}
    a_t = \begin{cases} 
        \text{random action}, & \text{if } \epsilon > \text{random value} \\
        \arg\max_a Q(s_t, a), & \text{otherwise}
    \end{cases}
\end{equation}

where $\epsilon$ decays according to:

\begin{equation}
    \epsilon = \max(\epsilon_{\text{min}}, \epsilon \times \epsilon_{\text{decay}})
\end{equation}

The Q-values are updated using the Bellman equation:

\begin{equation}
    Q(s_t, a_t) = R_t + \gamma \max_{a'} Q(s_{t+1}, a')
\end{equation}

where $R_t$ represents the reward at time step $t$, $\gamma$ is the discount factor, and $s_{t+1}$ is the next state after taking action $a_t$. The loss function for training the Q-network is defined as:

\begin{equation}
    L(Q) = \frac{1}{N} \sum_{i=1}^{N} \left(Q(s_t, a_t) - \left[ R_t + \gamma \max_{a'} Q(s_{t+1}, a')\right]\right)^2
\end{equation}

where $N$ is the batch size during training. The Q-network consists of fully connected layers with the following architecture:

\begin{itemize}
    \item Layer 1: Dense with 64 units, ReLU activation
    \item Layer 2: Dense with 32 units, ReLU activation
    \item Layer 3: Dense with 16 units, ReLU activation
    \item Layer 4: Dense with 8 units, ReLU activation
    \item Output layer: Dense with $\text{action\_dim}$ units (linear activation)
\end{itemize}

Each layer $l$ computes the output as:

\begin{equation}
    h^l = \text{ReLU}(W^l h^{l-1} + b^l)
\end{equation}

where $W^l$ and $b^l$ denote the weights and biases of layer $l$. As training progresses, $\epsilon$ decreases, reducing exploration and increasing exploitation. Through DQN, agents develop policies that optimise cumulative reward by selecting regions with favourable AQI. The procedure is outlined in Algorithm~\ref{alg:dqn}.

\begin{algorithm}[htbp]
    \caption{DQN training in the AQI-integrated world-grounded environment}
    \label{alg:dqn}
    \begin{algorithmic}[1]
        \State \textbf{Input:} 
        \State \quad State size $(\text{latitude}, \text{longitude}, \text{AQI})$
        \State \quad Action size $(\text{north}, \text{south}, \text{east}, \text{west})$
        \State \textbf{Hyperparameters:} 
        \State \quad Discount factor $(\gamma)$, Learning Rate $(\alpha)$
        \State \quad Exploration Rate $(\epsilon, \epsilon_{\text{min}}, \epsilon_{\text{decay}})$
        \State \quad Batch Size $(N)$
        \State \textbf{Output:}
        \State \quad Optimal policy for agents
        \State \quad Total reward accumulated
        \State \quad Traversal coordinates
        
        \Function{InitializeEnvironment}{}
            \State Initialize replay memory $D$ to store past experiences
            \State Initialize primary network $Q$ and target network $Q'$ with random weights
            \State Set environment with AQI, bird call data, and state boundaries
        \EndFunction
        
        \Function{TrainDQN}{}
            \For{each episode}
                \State Reset environment, get initial state $s_1$
                \For{each step $t$}
                    \State Select action $a_t$ using epsilon-greedy policy
                    \State Execute action, observe reward $R_t$ and next state $s_{t+1}$
                    \State Store transition $(s_t, a_t, R_t, s_{t+1})$ in $D$
                    \State Sample mini-batch from $D$
                    \State Compute target values based on Bellman equation
                    \State Train $Q$ by minimizing loss function $L(Q)$
                    \State Periodically update target network $Q' \leftarrow Q$
                    \State Update state $s_t \leftarrow s_{t+1}$
                    \State Decay $\epsilon$
                \EndFor
            \EndFor
        \EndFunction
    \end{algorithmic}
\end{algorithm}

The computational demands of deep reinforcement learning (DRL) pose challenges for scalability and practical use in large-scale environmental modelling. Training a DQN requires substantial computation because of the state space, the reward structure and iterative learning updates. As the number of agents grows---each initialised from a bird record and navigating the AQI-mapped environment---memory, processing and storage requirements grow accordingly. Applications that need fast inference may also be limited by networks with high computational overhead. Cloud-based and parallelised training can mitigate these limits, but deployment in resource-constrained settings remains a challenge; network pruning, quantisation and adaptive sampling are possible optimisations.

With DQN, agents learn navigation strategies driven by AQI, which allows the framework to simulate movement patterns that respond to air quality and geographic constraints. The integration of measured AQI data with reinforcement learning makes these agent--environment interactions explicit and open to analysis.

\subsection{Multi-Instance Runtime Integration}

A key aspect of the framework is the integration of the custom environment with the DQN model. Training begins by loading the data sources---bird-recording metadata, the AQI grid matrix and the states or sub-territories of the study region---so that the simulation is grounded in the real geography and air quality of the region. Agent movement is shaped by variations across environmental zones as given by the AQI data: the agents adjust their moves north, south, east or west in response to air-quality conditions, in a manner loosely analogous to birds adjusting their flight in response to environmental change.

Learning is organised into structured episodes that together involve more than 400 agents, which optimises computational resource allocation and streamlines data processing. Each agent derives feedback from the rewards obtained for its actions, giving an efficient mechanism for performance evaluation. Analogous to observing a large flock during a single migration period, this design gives a collective snapshot of agent behaviour, and analysing how agents modify their behaviour in response to environmental variation provides insight into their learning dynamics. Agents or timesteps that share a training run should not, however, be treated as independent replications.

To mitigate overfitting, the behaviour of the DQN model is regulated. The learned weights are monitored, and when an agent exhibits repetitive behaviour, the agent and the DQN model are reset to previously successful weight configurations. This prevents agents from becoming trapped in rigid action sequences and encourages adaptive learning. In addition, agents that exhibit repetitive actions are penalised with a reward of $-10$, and agents that consistently fail to perform are excluded from further training, which encourages diverse strategies. These mechanisms are training heuristics rather than established ecological processes; their effect on generalisation requires ablation, and agent removal changes the evaluated population, which must be considered when interpreting aggregate metrics.

The resulting runtime captures agent adaptability and exploratory tendencies while proactively addressing overfitting. The full procedure is given in Algorithm~\ref{alg:runtime_execution}.

\begin{algorithm}[htbp]
    \caption{Multi-instance runtime execution}
    \label{alg:runtime_execution}
    \begin{algorithmic}[1]
        \State \textbf{Input:} 
        \State \quad Preprocessed bird-recording data from Xeno-canto
        \State \quad State data, AQI matrix data
        \State \quad State and action dimensions
        \State \quad Learning rate, discount factor $(\gamma)$
        \State \quad Exploration rate $(\epsilon)$, decay factor $(\epsilon_{\text{decay}})$, minimum exploration rate $(\epsilon_{\text{min}})$
        \State \textbf{Output:} 
        \State \quad Total reward per agent
        \State \quad Trained DQN weights
        \State \quad Traversal coordinates
        
        \Function{InitializeEnvironment}{}
            \State Initialize custom environment with input data
            \State Initialize DQN agent with hyperparameters
            \State Set number of episodes
            \State Initialize lists to store episode rewards, paths, model weights, and epsilon values
        \EndFunction
        
        \Function{TrainAgents}{}
            \For{each episode}
                \State Reset environment to initial state
                \State Initialize variables to track episode rewards and paths
                \While{episode is not finished}
                    \State Select action using epsilon-greedy policy
                    \State Execute action, observe next state, reward, and done status
                    \State Train agent with transition $(s_t, a_t, r_t, s_{t+1})$
                    \State Update state and cumulative episode reward
                    \If{length of action list $>$ 70 and majority of actions are repetitive}
                        \State Reset model weights to last saved state
                        \State Reset $\epsilon$ value to last saved state
                        \State Set flag condition as true
                    \EndIf
                    \If{actions persist for extended duration}
                        \State Display status (reward total, actions, location data, etc.)
                        \State Append total reward to episode rewards list
                        \State Prepare environment for next agent
                        \State Store traversal data into paths list
                        \State Empty action list and continue execution
                    \EndIf
                    \If{done condition is true}
                        \State Display episode status
                        \State Store episodic data
                        \If{flag condition is true}
                            \State Store trained weights and $\epsilon$ value
                        \EndIf
                    \EndIf
                \EndWhile
                \State Append episode rewards and status details to lists
            \EndFor
        \EndFunction
    \end{algorithmic}
\end{algorithm}

\subsection{Bird Occurrence Data and AQI Integration}

Bird data were sourced from Xeno-canto \cite{Xeno-canto}, a widely used platform hosting one of the most comprehensive collections of bird vocalisations and their metadata. India was selected as the study region because of its heterogeneous AQI distribution and rich avian biodiversity. The dataset was obtained through API requests to the Xeno-canto database, extracting latitude, longitude and altitude. The AQI at each recording location was computed with the Ozon3 library from its latitude and longitude. The final dataset comprises 479 instances, each representing a bird recording occurrence. These coordinates allow the recordings and their AQI values to be mapped, giving the spatial distribution of the recordings across India and the air-quality conditions at those locations.

These records do not provide individually linked, time-ordered tracks: the instances must not be interpreted as tracked birds or independent experimental replications, and the generated paths are simulated trajectories rather than reconstructions of observed migration. Recording effort, accessibility and species detectability may influence the sampled locations, and spatial association with the AQI matrix does not establish contemporaneous exposure unless recording dates and pollution timestamps are matched.

\begin{figure}[htbp]
    \centering
    \includegraphics[width=0.62\textwidth]{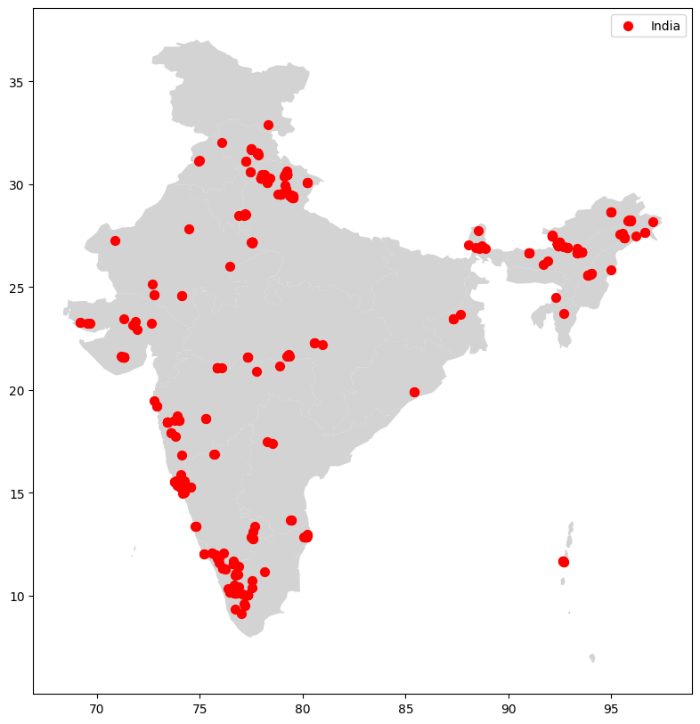}
    \caption{Bird-recording locations from the Xeno-canto dataset (red dots) overlaid on the shapefile of India.}
    \label{fig:j2}
\end{figure}

\begin{figure}[htbp]
    \centering
    \includegraphics[width=0.62\textwidth]{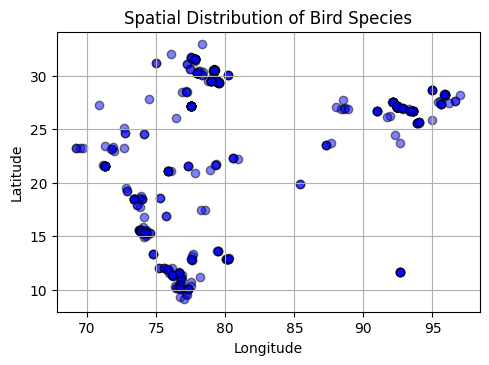}
    \caption{Spatial distribution of the bird-recording locations by latitude and longitude.}
    \label{fig:j3}
\end{figure}

\begin{figure}[htbp]
    \centering
    \begin{minipage}[t]{0.47\textwidth}\centering
    \includegraphics[width=\linewidth]{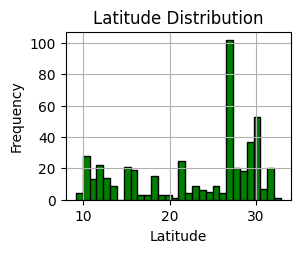}
    \caption{Histogram of recording latitudes.}\label{fig:j4}
    \end{minipage}\hfill
    \begin{minipage}[t]{0.47\textwidth}\centering
    \includegraphics[width=\linewidth]{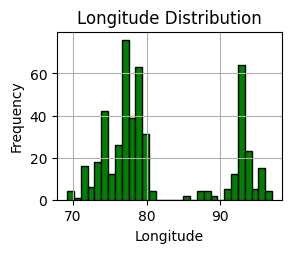}
    \caption{Histogram of recording longitudes.}\label{fig:j5}
    \end{minipage}
\end{figure}

The dataset contains geographical coordinates (latitude and longitude) for 478 locations across India. The descriptive statistics for latitude include a mean of 22.992704 degrees and a standard deviation of 7.135504 degrees, with values ranging from a minimum of 9.136100 degrees to a maximum of 32.911200 degrees. The median latitude is 26.767300 degrees, with the 25th and 75th percentiles at 15.544200 and 28.633350 degrees, respectively. For longitude, the mean is 81.401885 degrees with a standard deviation of 7.674811. Longitude values range from 69.203600 degrees to 97.021000 degrees, with a median of 78.369100 degrees, while the 25th and 75th percentiles are at 76.250775 and 92.106519 degrees, respectively. The dataset includes 226 unique locations corresponding to sightings of 41 bird species. Figure~\ref{fig:j2} presents the locations of these observations, while Figure~\ref{fig:j3} illustrates their spatial distribution across India, with blue dots representing bird sightings. Figures~\ref{fig:j4} and~\ref{fig:j5} display the distributions of latitude and longitude values, respectively, providing insights into the geographical spread of observations.

\begin{figure}[htbp]
    \centering
    \includegraphics[width=0.7\textwidth]{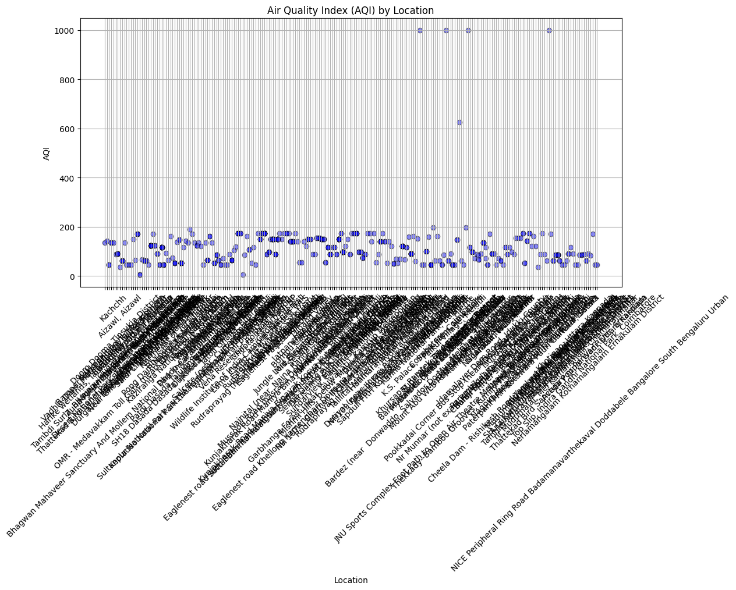}
    \caption{AQI values at each of the 226 unique recording locations.}
    \label{fig:j6}
\end{figure}

\begin{figure}[htbp]
    \centering
    \includegraphics[width=0.75\textwidth]{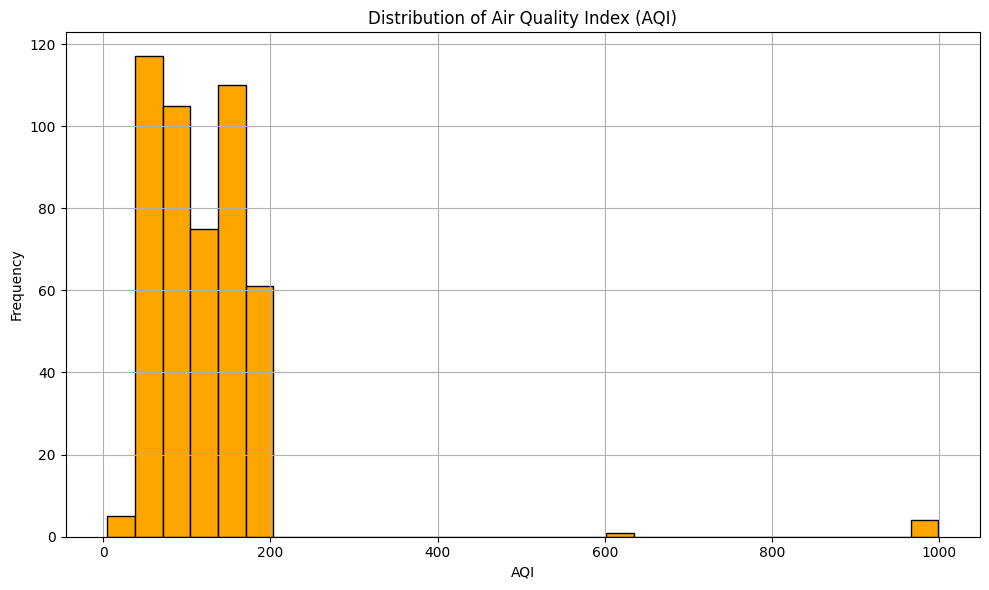}
    \caption{Histogram of AQI values across the 226 unique recording locations.}
    \label{fig:j7}
\end{figure}

In terms of AQI, the dataset includes 478 observations with a mean AQI of 111.57 and a standard deviation of 49.84. The AQI values range from 5 to 999, with quartile values at 25\% (Q1) = 71, 50\% (median, Q2) = 118, and 75\% (Q3) = 151. Figure~\ref{fig:j6} visualizes AQI values across study locations, with the x-axis representing location indices and the y-axis denoting AQI values within the range of 5 to 999. This plot highlights spatial variability in air quality across different study regions. Figure~\ref{fig:j7} provides a histogram of AQI frequency distribution across 226 locations, with AQI values plotted along the x-axis and their corresponding frequency along the y-axis, offering an overview of air quality variability in the dataset.

\begin{figure}[htbp]
    \centering
    \includegraphics[width=0.62\textwidth]{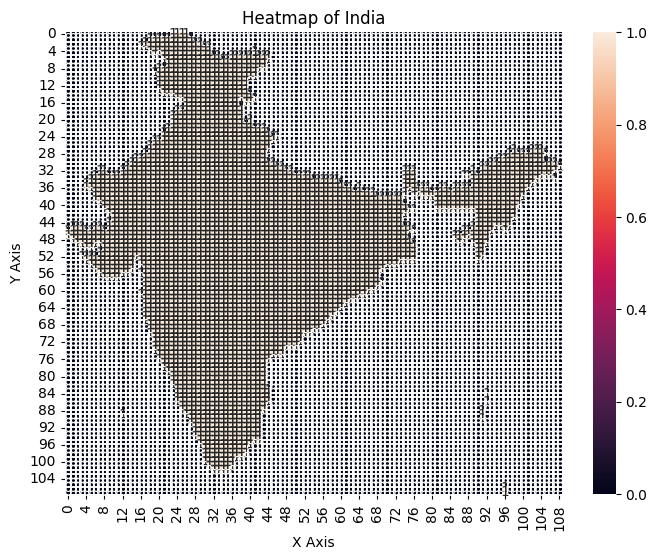}
    \caption{State-wise binary mask of India used to define valid cells of the environment and to update the AQI matrix.}
    \label{fig:j8}
\end{figure}

\begin{figure}[htbp]
    \centering
    \includegraphics[width=0.62\textwidth]{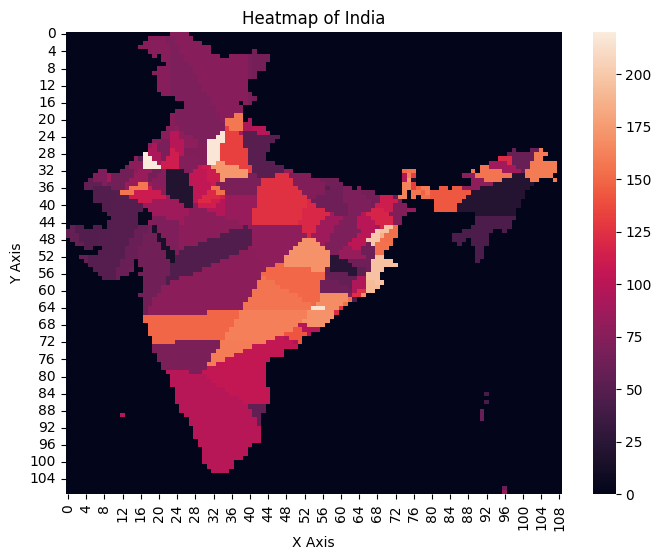}
    \caption{Heatmap of the precomputed AQI matrix over India, the environmental feedback layer of the world-grounded environment.}
    \label{fig:j9}
\end{figure}

Figures~\ref{fig:j8} and~\ref{fig:j9} add state boundaries and the AQI heatmap of India, respectively, both produced by the framework, which integrates air-quality data with the geographic grid. Figure~\ref{fig:j8} shows the state-wise mask, and Figure~\ref{fig:j9} shows the spatial distribution of AQI across the country, which forms the environmental layer of the simulation. The shapefile of India was sourced from a government data portal \cite{DataGov2023}, with remote-sensing resources~\cite{india_remote_sensing} providing state and union-territory delineations. Shapefiles for other countries can be obtained from sources such as OpenStreetMap, and the framework can be connected to other geographic data sources, which makes it applicable to other regions.

\subsection{Scope of the Simulation}

The framework is a computational model of pollution-aware navigation, and several of its components are modelling choices whose scope should be explicit. Bird movement is represented as navigation over the AQI-integrated grid with four discrete actions of fixed $0.27^\circ$ shifts in latitude or longitude; the observation includes geographic position and AQI, so that decisions are driven by both spatial and environmental signals. The fixed AQI matrix represents spatial heterogeneity, not temporal pollution dynamics. Grid steps are computational transitions without a calibrated flight duration, and fixed angular increments do not represent equal physical distances across latitudes. The study therefore examines an explicit navigation objective in a world-grounded environment without claiming that birds perceive AQI or optimise the same reward.

\section{Experiments and Results}\label{sec:results}

This section presents the deployment of the framework on India, with all mappings integrated and bird propagation simulated across the region. Bird data for India from Xeno-canto were used to initialise the agents. Graphical analyses and quantitative metrics are presented to assess the behaviour and learning of the framework, covering how agent movement responds to air quality and geographic constraints and how the agents' decision-making evolves during training.

\subsection{Experimental Setup}

An empirical analysis of the proposed methodology was conducted by integrating reinforcement learning (RL) frameworks and simulations, which were implemented using Python within a Windows 10 environment. The computational setup consisted of a system equipped with 8GB of RAM and a Ryzen 5 5000 series CPU. All experiments were executed using Python version 3.10, leveraging Google Collaboratory for computational resources. This experimental configuration provided the requisite environment for running RL simulations and systematically evaluating the performance of the proposed approach.

\subsection{Evaluation Metrics}\label{sec:metrics}

We evaluate five simulation metrics. They quantify the specified simulation objective rather than biological fitness.

\paragraph{Cumulative reward.}
\begin{equation}
R_{\text{cumulative}} = \sum_{t=1}^{T} r_t
\end{equation}
where $r_t$ is the reward at timestep $t$ and $T$ is the total number of timesteps in an episode.

\paragraph{Reward change rate.}
\begin{equation}
\Delta R_t = r_t - r_{t-1}, \quad \forall t \in \{2,3,\dots,T\}
\end{equation}
which describes consecutive reward differences.

\paragraph{Action diversity (entropy).}
\begin{equation}
D_{\text{action}} = - \sum_{a \in A} p(a)\log p(a)
\end{equation}
where $A$ is the set of possible actions and $p(a)=n(a)/N$, with $n(a)$ the number of times action $a$ is selected and $N$ the total number of actions. Its exponential moving average is
\begin{equation}
D_{\text{action,EMA}}(t) = \alpha D_{\text{action}}(t) + (1-\alpha) D_{\text{action,EMA}}(t-1),
\end{equation}
with smoothing factor $0<\alpha\leq1$. Entropy measures action frequencies and does not by itself demonstrate adaptive behaviour.

\paragraph{Total distance travelled.}
\begin{equation}
D_{\text{travelled}} = \sum_{t=2}^{T} \left\| s_t - s_{t-1} \right\|
\end{equation}
where $s_t$ denotes geographic position only, excluding AQI and state code, and $\|\cdot\|$ is the Euclidean distance in the coordinate representation. Computed from latitude and longitude, this is angular displacement rather than physical flight distance, which would require a geodesic calculation.

\paragraph{Step efficiency.}
\begin{equation}
E_{\text{step}} = \frac{R_{\text{cumulative}}}{T}
\end{equation}
the mean reward per timestep.

\paragraph{Mean absolute error.} To monitor learning, the mean absolute error (MAE) of the agent model is tracked against the steps taken by each agent (Section~\ref{sec:loss}).

\subsection{Multi-Instance DQN Environment Analysis}

This section analyses the aggregated rewards and traversal paths of the agents after all DQN agents initialised from the Xeno-canto dataset have been executed. Each agent corresponds to a data record with its own initial state of longitude, latitude and AQI. Throughout the simulation, whether an action comes from the DQN's prediction or from exploration, the cumulative reward of each successful step is logged, as described in the methodology. For spatial analysis and visualisation, the coordinates of the agents' movement across India are also recorded, which is needed to understand collective agent behaviour under the varying AQI of different locations.

Beyond reward analysis, the learning error of the agents is evaluated and visualised to monitor learning efficiency and adaptation, which helps assess the convergence behaviour of the framework. In addition, a comparative analysis of different RL algorithms is carried out, covering reward accumulation, traversal and decision-making behaviour, to assess how the proposed approach behaves relative to standard alternatives in the same world-grounded environment.

\subsubsection{Performance Analysis of Agents}

\begin{figure}[htbp]
    \centering
    \includegraphics[width=0.75\textwidth]{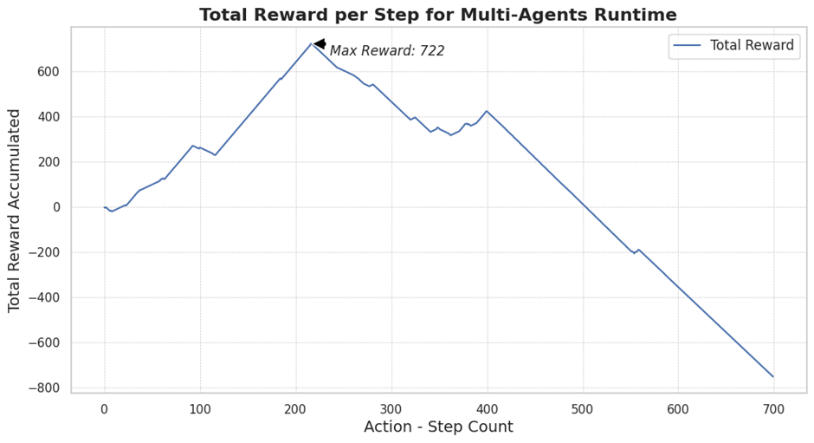}
    \caption{Cumulative reward collected at each step by the DQN agents, peaking at 722.}
    \label{fig:j10}
\end{figure}

As outlined above, the behavioural and learning trends of the DQN model are examined using reward data together with the states and actions of the agents. The reward plot in Figure~\ref{fig:j10} shows the cumulative reward accumulated at each action step. The trend first increases, reaching a peak reward of 722, as agents find regions with valid AQI values that satisfy the reward conditions. A subsequent decline in total reward follows from environmental and model constraints, including boundary violations, cells with zero AQI and AQI comparisons outside the acceptance band. These constraints are part of the environment design and keep agent movement within the world-grounded region.

The key environmental constraints incorporated into the framework are:

\begin{itemize}
    \item \textbf{Boundary violations:} A penalty of $-5$ is imposed and the agent's position is reset to its original state. These violations represent natural and geopolitical boundaries and prevent movement outside the modelled region, keeping trajectories within the land-masked grid.
    \item \textbf{Zero-AQI cells:} A penalty of $-4$ is applied when agents enter cells with zero AQI, which correspond to locations such as oceans or cells outside the study scope. The penalty steers agents back toward valid land cells.
    \item \textbf{AQI outside the acceptance band:} Agents are penalised when they reach cells whose AQI falls outside the acceptance band defined relative to their reference AQI (Equation~\eqref{eq:rband}). This keeps exploration within conditions similar to the agent's starting environment, which the framework uses as a proxy for ecological tolerance.
\end{itemize}

Integrating these constraints into the RL framework keeps agent behaviour consistent with the modelled geography and the specified objective. The penalties also improve learning efficiency by directing agents toward valid regions and reduce overfitting to specific locations.

The reward plot of the DQN model gives insight into agent--environment interaction. From step 0 to step 200, total reward increases consistently, indicating effective adaptation. The peak reward of 722 around step 200 corresponds to a phase in which agents successfully apply learned strategies. The subsequent decline corresponds to agents moving into new territory and incurring penalties that reduce cumulative reward. This balance between exploration and exploitation is a characteristic of the RL framework.

The reduction in reward after the peak plays a role in learning: by penalising agents in less suitable regions, the model prevents over-reliance on known areas and encourages agents to experience a broader range of environmental conditions. The reward dynamics show how the model evaluates cells under varying AQI, which is relevant for using the framework to study movement within real-world environmental constraints.

To further assess the framework, a comparative analysis is conducted across different RL algorithms in the same environment. The resulting metrics provide a baseline for future RL models in world-grounded ecological simulation.

\subsubsection{Comparison with Classical RL Methods}

\begin{figure}[htbp]
    \centering
    \includegraphics[width=0.75\textwidth]{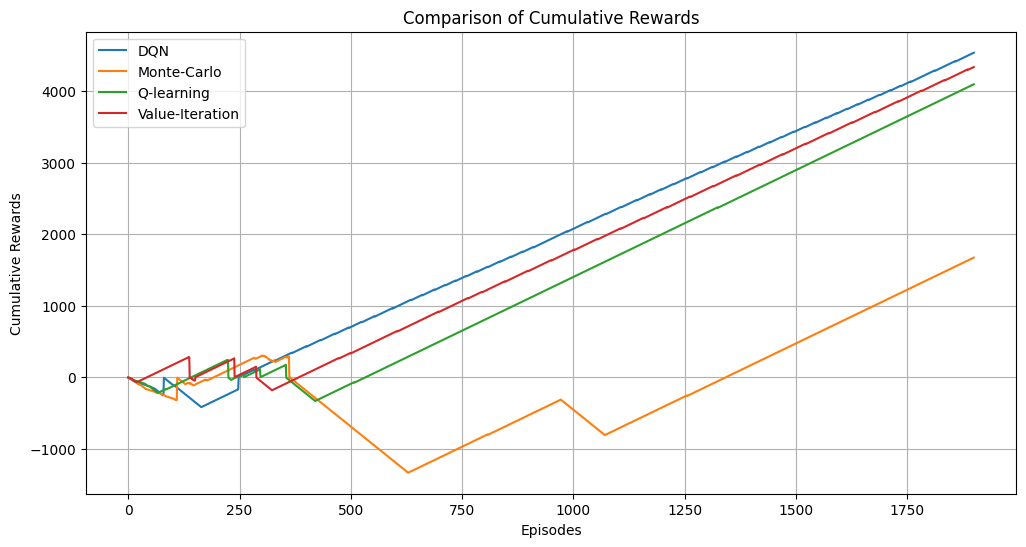}
    \caption{Cumulative reward per episode for DQN, Monte Carlo, Q-Learning and Value Iteration in the world-grounded environment.}
    \label{fig:j11}
\end{figure}

Figure~\ref{fig:j11} presents a detailed comparative analysis of the performance of the four agents within the AQI-integrated environment. This analysis explores the underlying mechanisms governing agent behavior, focusing on how each approach influences performance within this dynamic and complex setting. The Deep Q-Network (DQN) demonstrates superior performance among all methods, achieving the highest cumulative rewards as episodes progress. Its trajectory is characterized by a steady and steep upward trend, reflecting its capacity to effectively learn optimal policies. After approximately 250 episodes, DQN's performance distinctly surpasses other methods. While initial growth is gradual, the model stabilizes quickly and exhibits a steep increase in cumulative rewards, indicative of efficient learning through experience replay and neural network function approximation. The sharp improvement beyond 250 episodes demonstrates the model’s enhanced capacity to generalize across the state-action space. Unlike other methods, DQN does not exhibit oscillations or slow convergence, maintaining a consistently strong trajectory across episodes. This superior performance is attributed to DQN's use of deep neural networks to approximate the Q-value function and the implementation of experience replay to stabilize learning. These mechanisms enable effective generalization in complex, high-dimensional state spaces.

Conversely, the Monte Carlo method exhibits the weakest performance, as reflected in its significantly lower cumulative rewards and unstable trajectory in earlier episodes. Between 0 and 500 episodes, Monte Carlo demonstrates pronounced instability, with cumulative rewards declining into negative values and recovering intermittently. This behavior is attributed to its reliance on high-variance updates based on complete episode returns. Recovery begins around 500 episodes, where a noticeable upward trend is observed; however, growth remains slower relative to other methods. The instability of Monte Carlo in early episodes underscores its inefficiency in environments that require incremental updates, and its delayed recovery suggests a prolonged convergence period for optimal policies.

Q-Learning and Value Iteration exhibit intermediate performance levels, with both demonstrating steady and consistent increases in cumulative rewards. Q-Learning, despite outperforming Monte Carlo, lags slightly behind Value Iteration and DQN. This performance is attributed to its reliance on tabular Q-value updates, which do not scale as efficiently as the function approximation techniques employed in DQN. Initially, cumulative rewards exhibit gradual growth, reflecting the iterative nature of tabular updates. Between 200 and 500 episodes, a pronounced upward trend is observed as the algorithm refines its policy. However, its growth rate eventually stabilizes, with DQN and Value Iteration surpassing it in later episodes. This plateau effect results from Q-Learning’s limited scalability in high-dimensional state spaces without function approximation.

Value Iteration benefits from its model-based approach, allowing for more efficient convergence compared to Q-Learning. However, its inability to match the scalability and representational power of DQN becomes apparent as episodes increase, with DQN ultimately achieving superior cumulative rewards. Initially, Value Iteration exhibits strong performance, demonstrating steady growth from the outset of the simulation. Between 0 and 250 episodes, its trajectory slightly exceeds that of Q-Learning, indicating faster convergence during early training phases. This early advantage stems from its model-based nature, which enables efficient computation of optimal policies using the environment’s transition dynamics. However, beyond 500 episodes, growth decelerates, and cumulative rewards are ultimately surpassed by DQN. This slowdown suggests that while Value Iteration achieves rapid initial convergence, it lacks the scalability and flexibility of neural network-based approaches in complex environments.

These performance trends highlight the diverse learning dynamics of the evaluated algorithms. DQN excels in both stability and scalability, demonstrating superior performance beyond 250 episodes. Value Iteration and Q-Learning exhibit strong early-phase learning but experience slower long-term growth. Monte Carlo’s poor initial performance and delayed recovery further emphasize its limitations in environments requiring incremental and stable policy updates.

\begin{figure}[htbp]
    \centering
    \includegraphics[width=0.75\textwidth]{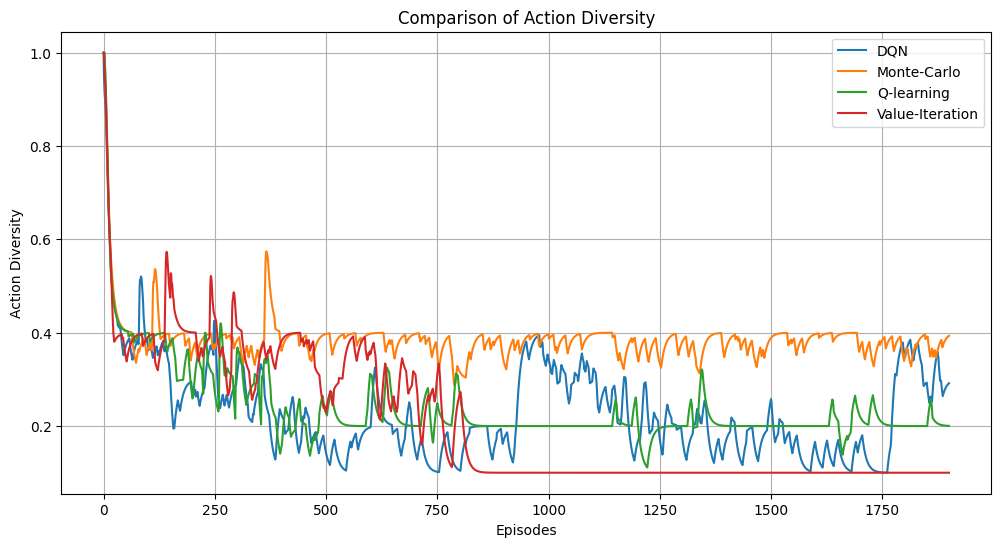}
    \caption{Action diversity (entropy-based) per episode for the four algorithms.}
    \label{fig:j12}
\end{figure}

Figure~\ref{fig:j12} illustrates the evolution of action diversity across training episodes for four reinforcement learning algorithms. Action diversity, which represents the variability in selected actions, provides insights into an agent’s balance between exploration and exploitation and the refinement of its policy over time. The analysis reveals significant differences in algorithmic behavior within the AQI-integrated environment, emphasizing how their respective exploration strategies influence learning dynamics.

DQN exhibits high action diversity at the beginning of training, indicative of its initial exploratory phase, which is guided by the epsilon-greedy exploration strategy. A sharp decline in diversity is observed within the first 100 episodes, signaling a rapid transition toward policy exploitation. However, complete elimination of variability does not occur, as action diversity stabilizes at moderate levels throughout the remainder of training. This behavior reflects the dynamic nature of DQN, where the gradual decay of the epsilon parameter ensures that a nonzero probability of exploration is maintained. The residual variability allows the model to adapt to environmental changes and mitigate the risk of convergence to suboptimal policies. Periodic fluctuations in later episodes suggest that strategic exploration continues, which is particularly advantageous in environments characterized by complex dynamics or high-dimensional state spaces. This ability to balance exploration and exploitation contributes to the robustness of DQN, as evidenced by its superior cumulative rewards (Figure~\ref{fig:j11}). The moderate yet consistent action diversity underscores the flexibility of DQN’s neural network-based architecture, facilitating generalization across a broad range of state-action pairs.

Monte Carlo exhibits the highest and most sustained action diversity among all evaluated algorithms. In the initial episodes, significant fluctuations in diversity are observed, with pronounced peaks and troughs indicating instability in policy updates. This behavior is expected, as Monte Carlo relies on episodic returns, which are inherently high in variance due to their dependence on complete episode outcomes. After 500 episodes, action diversity stabilizes but remains consistently high, often exceeding 0.4 throughout training. The sustained variability indicates a strong emphasis on broad exploration of the state-action space, making Monte Carlo suitable for environments where exhaustive exploration is essential for identifying optimal policies. However, this extensive exploration comes at the cost of delayed convergence and reduced policy stability. The persistent high action diversity suggests difficulty in stabilizing a deterministic policy, particularly in dynamic or noisy environments. Consequently, Monte Carlo may exhibit suboptimal performance in scenarios that require rapid convergence or fine-tuned exploitation strategies.

Q-Learning demonstrates a steady decline in action diversity, with a distinct transition from exploration to exploitation occurring around 200–250 episodes. Initially, moderate variability is observed, reflecting the exploration phase driven by the epsilon-greedy strategy. As Q-values are iteratively refined, action diversity decreases, eventually stabilizing at approximately 0.2. This stabilization indicates a transition toward a predominantly deterministic policy once optimal Q-values for state-action pairs are established. The low action diversity following convergence highlights Q-Learning’s strength in policy exploitation, wherein actions with the highest Q-values are consistently selected. However, this deterministic behavior may limit adaptability, reducing the algorithm’s ability to respond to unforeseen changes or explore alternative policies. In environments with high-dimensional or partially observable state spaces, this reduction in exploration could lead to suboptimal performance. Nonetheless, the structured and predictable reduction in action diversity reflects the stability and systematic nature of Q-Learning’s learning process.

Value Iteration maintains minimal action diversity throughout the training process, stabilizing near zero after the initial episodes. This behavior is a direct consequence of the algorithm’s reliance on a fixed model of the environment’s transition dynamics. Unlike other algorithms, Value Iteration does not incorporate stochastic exploration mechanisms such as epsilon-greedy strategies. Instead, it computes a deterministic policy by iteratively solving the Bellman equation for each state-action pair. While this deterministic approach facilitates rapid and reliable convergence, it severely restricts the algorithm’s ability to explore alternative actions or adapt to dynamic environments. The absence of variability renders Value Iteration well-suited for tasks with fully known and static dynamics but poorly equipped for problems requiring flexible adaptation or learning under uncertainty. The consistently low action diversity underscores the rigidity of the algorithm, which, although computationally efficient in predictable settings, limits its broader applicability.

The action diversity analysis provides critical insights into the exploration strategies and learning dynamics of these reinforcement learning algorithms. DQN emerges as the most versatile approach, effectively balancing exploration and exploitation to adapt to complex environments. Monte Carlo exhibits the highest exploratory tendencies but struggles with policy stability and convergence. Q-Learning and Value Iteration demonstrate increasingly deterministic behavior, with the former achieving gradual convergence and the latter prioritizing rapid stabilization. These findings highlight the trade-offs between exploration and exploitation across different RL algorithms, influencing their suitability for various ecological modeling tasks.

\begin{figure}[htbp]
    \centering
    \includegraphics[width=0.75\textwidth]{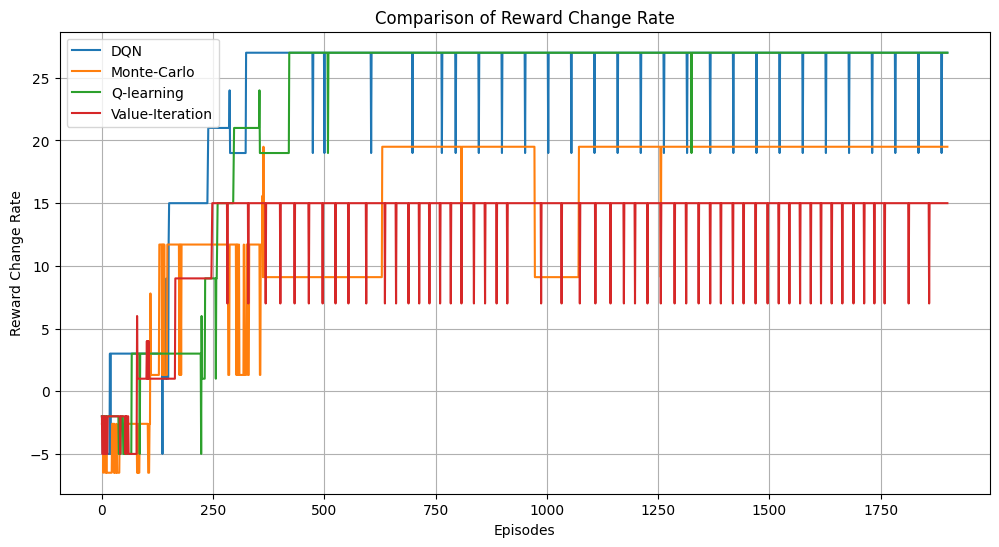}
    \caption{Reward change rate per episode for the four algorithms.}
    \label{fig:j13}
\end{figure}

Figure~\ref{fig:j13} illustrates the evolution of cumulative reward changes per episode. The vertical axis represents the reward change rate, while the horizontal axis denotes the number of episodes. Each trajectory corresponds to the reward change rate of an individual algorithm over time. During the initial learning phase, both DQN and Q-learning exhibit rapid learning, achieving high reward change rates early in training.

The performance trajectory of the Deep Q-Network (DQN) is characterized by an initial phase of rapid improvement followed by stabilization. Within the first 250 episodes, a steep increase in the reward change rate is observed, indicating the algorithm’s ability to quickly adapt and optimize its policy. This rapid ascent is facilitated by DQN’s deep learning architecture, which enables effective approximation of Q-values and efficient generalization of state-action value mappings. The $\epsilon$-greedy exploration strategy ensures adequate exploration, while the neural network rapidly prioritizes actions that yield higher rewards. The aggressive learning phase underscores DQN’s strength in navigating high-dimensional and complex environments. As training progresses, the reward change rate stabilizes at a consistently high level, indicating convergence toward an optimal or near-optimal policy. The minimal fluctuations observed in later episodes highlight the robustness of DQN, demonstrating its ability to maintain stable policies even in dynamic environments. The sustained high performance suggests that the learned policies are not only optimal but also resilient to perturbations in input conditions.

Q-learning follows a similar trend, with a rapid increase in the reward change rate during the early episodes. Its off-policy nature accelerates the learning process by allowing updates from actions outside the current policy. The deterministic update rule, which incorporates maximum future rewards, further contributes to the fast learning rate. Compared to Monte Carlo, Q-learning stabilizes earlier and exhibits reduced fluctuation, indicating a more stable learning process. The lower variability suggests robustness in policy convergence, despite the algorithm’s known susceptibility to overestimation bias in Q-values. The overall consistency in performance highlights Q-learning’s ability to effectively balance exploration and exploitation, making it a reliable approach for reinforcement learning tasks.

Monte Carlo and Value Iteration also demonstrate effective learning but require a longer stabilization period due to differences in update mechanisms and algorithmic sensitivities. Between 250 and 750 episodes, Monte Carlo exhibits significant fluctuations in the reward change rate, reflecting its inherent instability arising from the high variance in episodic returns. Unlike Q-learning and DQN, which update policies more frequently and smooth out learning trajectories, Monte Carlo’s reliance on complete episodes results in volatile changes when policies are adjusted based on cumulative episode outcomes. Beyond 750 episodes, the reward change rate begins to stabilize, though variability remains present. This suggests gradual convergence toward an optimal policy while highlighting the method’s sensitivity to episodic return variance and the initial state distribution. The learning curve of Monte Carlo is less smooth compared to other algorithms, with higher variance and less frequent updates slowing down the learning process. However, its ability to capture long-term dependencies and delayed rewards makes it advantageous in tasks where these factors play a critical role.

Value Iteration exhibits a rapid increase in the reward change rate similar to DQN but with more pronounced fluctuations in the early episodes. This initial variability is attributed to the iterative nature of value updates, which can cause oscillations as the algorithm seeks to converge to the optimal value function. As training progresses, the reward change rate stabilizes, albeit with minor fluctuations. This indicates that while Value Iteration achieves high rewards early in training, maintaining consistent performance may be challenging. The observed fluctuations may result from the algorithm’s sensitivity to the precision of value updates and its deterministic policy structure. Despite these challenges, Value Iteration is theoretically guaranteed to converge to the optimal policy under certain conditions, making it well-suited for applications requiring optimality guarantees, such as planning and scheduling tasks.

In terms of overall performance, DQN achieves the highest and most stable reward change rate, making it the most effective algorithm in this comparison. Q-learning also performs well, striking a balance between rapid learning and stability. Monte Carlo’s performance, though variable, remains suitable for scenarios where episodic returns are significant. Value Iteration, while effective, may require additional tuning to achieve long-term stability comparable to DQN and Q-learning.

\begin{figure}[htbp]
    \centering
    \includegraphics[width=0.75\textwidth]{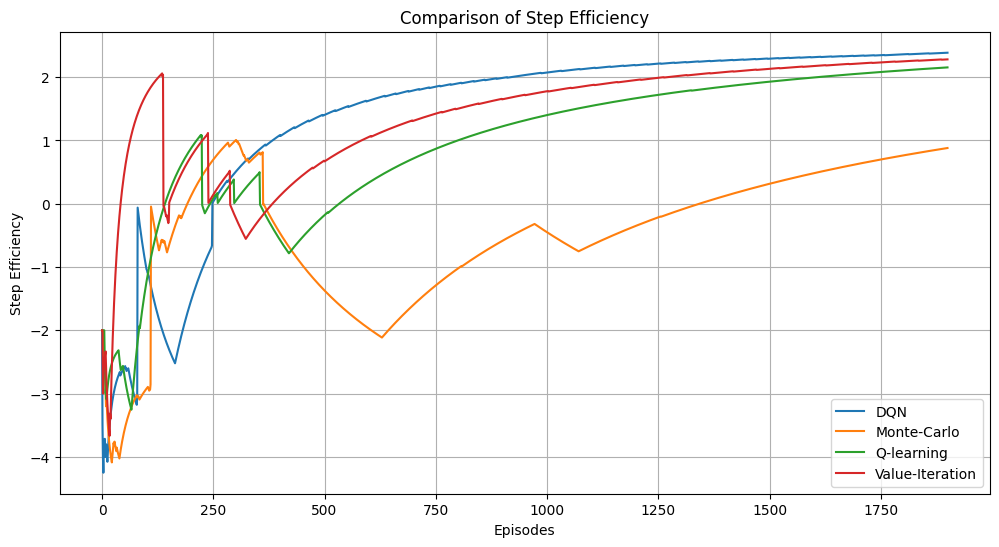}
    \caption{Step efficiency (reward per timestep) across episodes for the four algorithms.}
    \label{fig:j14}
\end{figure}

Figure~\ref{fig:j14} presents a comparative analysis of the four agents based on step-efficiency metrics across the training episodes. Step efficiency serves as a critical metric for evaluating how effectively an algorithm improves its policy per step or episode. A higher step efficiency indicates faster and more stable learning. 

DQN exhibits a robust and steady learning process. In the initial episodes, fluctuations in step efficiency are observed, starting at approximately -3 and gradually increasing as learning progresses. These early instabilities are mitigated through the use of neural networks for value function approximation and experience replay to stabilize learning. By the 250th episode, DQN demonstrates a consistent upward trajectory in efficiency, surpassing the 1.0 mark. This steady improvement persists through the mid-phase (250 to 750 episodes) and later phases (750 to 1750 episodes), with step efficiency approaching 2.0. DQN’s sustained increase in efficiency highlights its effectiveness in environments with large state-action spaces, where neural network generalization and experience replay contribute to balancing exploration and exploitation. This mechanism helps to break the correlation between consecutive updates, resulting in more stable and reliable learning. Over time, early fluctuations are further mitigated as the network refines its parameters. DQN’s capacity to continuously improve and maintain high step efficiency, particularly in later episodes, underscores its suitability for environments that require function approximation and stabilized learning processes.

Monte Carlo exhibits distinct learning characteristics due to its episodic nature. During the initial 250 episodes, significant variability in step efficiency is observed, fluctuating between -3 and values above 1.5. This high variance arises from the algorithm’s dependence on complete episodes for policy updates, leading to unstable early performance. As training progresses, a pronounced dip in efficiency is noted between 250 and 750 episodes, occasionally dropping below -2. This decline reflects the challenges posed by high variance in episodic returns, which complicates stable learning. Beyond 750 episodes, Monte Carlo gradually recovers and improves its step efficiency. However, despite showing a positive trend, its efficiency remains lower than that of other algorithms, indicating persistent sensitivity to episodic variance and less frequent policy updates. The results suggest that while Monte Carlo is effective in capturing long-term dependencies, its application is best suited for environments where episodes yield significant cumulative reward information, despite its slower convergence and higher variability.

Q-learning exhibits a balanced and steady learning curve. Initially, moderate fluctuations in step efficiency occur, reflecting the algorithm’s trade-off between exploration and exploitation. Step efficiency begins at approximately -3 but rapidly increases, stabilizing around 1.0 by the 250th episode. This stability is maintained through the mid-phase, with Q-learning displaying a smooth and consistent upward trajectory. In the later episodes, Q-learning converges towards a high step efficiency, slightly below that of DQN but still significant. The direct updates to Q-values allow Q-learning to refine its policy effectively, demonstrating robust learning dynamics. The results indicate that Q-learning performs well in environments with well-defined state-action spaces, where its incremental Q-value updates ensure stable and consistent improvements. Its effectiveness in such settings makes it a reliable choice for reinforcement learning tasks of moderate complexity.

Value Iteration, represented by the red line, achieves high and consistent step efficiency among the algorithms. From the initial episodes, Value Iteration exhibits a rapid increase in efficiency, quickly surpassing Q-learning and Monte Carlo. This early advantage is attributed to its reliance on a complete model of the environment, which allows for iterative computation of the optimal policy. By the 250th episode, Value Iteration attains efficiency values around 1.5, maintaining strong performance throughout the learning process. In the mid and later phases, step efficiency remains consistently around 2.0, demonstrating the algorithm’s stable learning dynamics. The results highlight Value Iteration’s advantages in environments where a full model is available, enabling efficient and effective policy optimization. However, despite its strong performance, Value Iteration falls slightly short of DQN, which benefits from neural network generalization and experience replay, resulting in marginally higher overall efficiency in complex environments. 

This comparative analysis highlights the distinct strengths and limitations of the four algorithms. Monte Carlo, while effective for episodic tasks with complete return information, suffers from high variance and slower convergence, limiting its efficiency in dynamic environments. DQN demonstrates robust performance, leveraging neural networks and experience replay to achieve steady learning and scalability in large and complex state-action spaces. Q-learning, with its direct Q-value updates, offers stability and moderate scalability, making it effective in environments with manageable complexity. Value Iteration emerges as the most step-efficient approach, excelling in model-based environments due to its iterative optimization capabilities. These insights underscore the importance of aligning algorithm selection with task characteristics and resource constraints to maximize learning efficiency and performance.

\begin{figure}[htbp]
    \centering
    \includegraphics[width=0.75\textwidth]{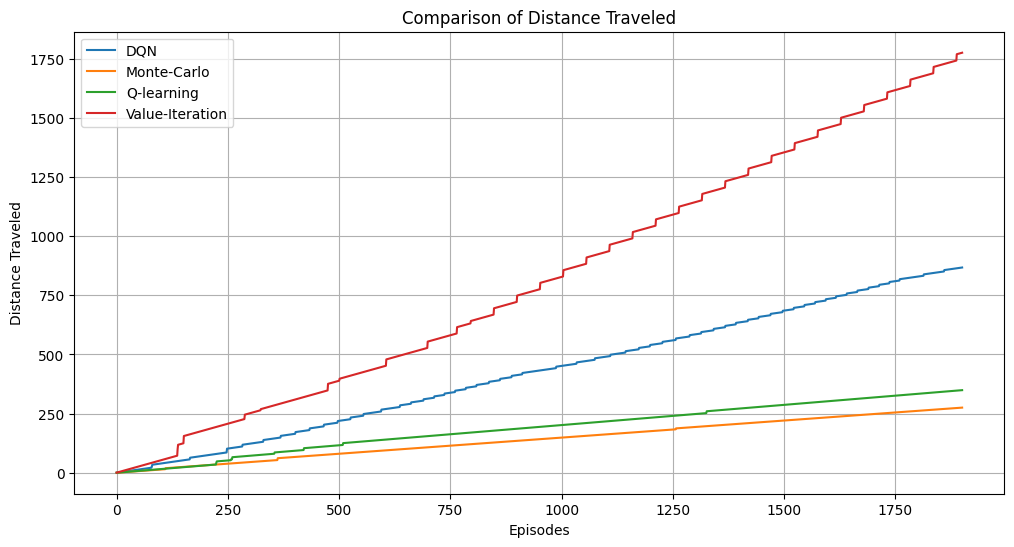}
    \caption{Cumulative distance travelled across episodes for the four algorithms (angular grid displacement).}
    \label{fig:j15}
\end{figure}

The total distance traveled by an agent serves as a crucial metric for evaluating its exploration behavior within an environment. By summing the Euclidean distances between successive positions, this metric provides insight into how effectively an agent navigates the state space to discover and refine its policy. Figure~\ref{fig:j15} presents the comparative analysis of the agents based on their cumulative distances traveled.

DQN exhibits steady growth in cumulative distance traveled, reflecting a well-balanced exploration strategy. During the initial episodes, the algorithm prioritizes exploration to gather diverse experiences, as evidenced by the gradual increase in distance. This exploration phase is facilitated by DQN's $\epsilon$-greedy policy and the use of experience replay, which ensures that the agent learns from varied transitions while mitigating the instability typically associated with online updates. Over time, the cumulative distance traveled increases at an approximately linear rate, indicating that DQN effectively transitions from exploration to exploitation, focusing on refining its policy to optimize performance. The substantial distance covered by DQN underscores its ability to systematically explore complex state spaces and adapt its policy to enhance efficiency.

Monte Carlo exhibits the slowest growth in cumulative distance traveled among the four algorithms, characterized by a nearly flat trajectory over the episodes. This behavior highlights the episodic nature of Monte Carlo, which updates its policy only at the conclusion of each episode. The dependence on complete returns introduces inefficiencies in exploration, as the agent may repeatedly follow suboptimal paths before adjusting its policy. Consequently, the limited distance covered suggests insufficient exploration of the state space, hindering the agent’s ability to discover improved policies. Although Monte Carlo remains theoretically robust in environments with well-defined episodes, its poor exploration efficiency in this context represents a fundamental limitation in tasks requiring dynamic and proactive policy refinement.

Q-learning demonstrates moderate exploration efficiency, as indicated by its cumulative distance traveled. In the early episodes, the agent covers a substantial area, reflecting an effective balance between exploration and exploitation. This efficiency arises from the temporal-difference learning framework, which enables incremental Q-value updates based on immediate feedback. However, as training progresses, the rate of growth in distance declines, suggesting a gradual shift toward exploitation. While this behavior signifies that Q-learning effectively consolidates its policy, the relatively lower cumulative distance compared to DQN and Value Iteration suggests limitations in exploring larger or more complex state spaces. Q-learning performs adequately in environments with simpler dynamics but may struggle to match the scalability of more advanced algorithms.

Value Iteration exhibits rapid and consistent growth in cumulative distance traveled, covering the largest total distance among the four algorithms. However, this extensive movement does not necessarily result in proportional performance improvements. The algorithm’s reliance on a complete model of the environment enables iterative computation of optimal policies, but its resulting behavior often entails exhaustive exploration of the state space, including regions that may not meaningfully contribute to policy refinement. This tendency toward over-exploration may highlight inefficiencies or suboptimal pathfinding, as the agent expends significant resources navigating areas of marginal value. While Value Iteration benefits from deterministic updates and comprehensive state space coverage, its excessive movement suggests a lack of focus in exploration compared to more adaptive algorithms such as DQN. As a result, although Value Iteration is thorough, it may not utilize its exploration efforts as efficiently as DQN, particularly in environments with complex or high-dimensional state spaces where targeted exploration is essential.

The comparative analysis reveals distinct exploration patterns and trade-offs among the four algorithms. DQN emerges as the most balanced and efficient in its exploration strategy, demonstrating a steady increase in cumulative distance while maintaining a strong focus on identifying high-value regions within the state space. This behavior highlights its capacity to adapt dynamically and refine its policy effectively in complex environments. Value Iteration, while achieving the greatest overall distance traveled, frequently engages in excessive and unfocused exploration, resulting in inefficiencies and suboptimal pathfinding in scenarios where targeted exploration is essential. This tendency underscores its reliance on exhaustive state space coverage, which may not always translate into optimal performance improvements. Q-learning exhibits moderate exploration efficiency, characterized by significant early improvements but a slower cumulative growth in later episodes. This pattern suggests a gradual shift toward exploitation, leading to a more conservative learning approach over time. Monte Carlo demonstrates the lowest exploration efficiency among the four algorithms, with limited cumulative distance traveled and a slow rate of improvement. This outcome reflects its episodic nature, which restricts its ability to adapt rapidly to environmental changes. 

These findings emphasize the importance of selecting an algorithm based on the specific requirements of the task. DQN is well suited for high-dimensional, complex environments due to its adaptive exploration-exploitation balance. Value Iteration performs optimally in fully modeled, structured tasks but may suffer from inefficient exploration. Q-learning provides a balance between simplicity and effectiveness in environments with manageable state spaces. Monte Carlo remains best suited for well-defined episodic environments where rapid adaptation is not a primary concern.

\subsubsection{Traversal Paths of the Agents}

Traversal coordinates recorded for each agent during runtime give insight into the behaviour and decision-making of the DQN model. The paths are computed on the India grid, which is aligned with the AQI matrix, so each cell traversed by an agent can be tracked. Analysing the paths shows the learning progression of the model, in which exploratory behaviour gradually gives way to exploitation as agents use acquired knowledge to navigate more efficiently---a fundamental characteristic of reinforcement learning.

Visualising the paths also shows where agents incur penalties, such as boundary violations or entry into unfavourable AQI regions, which supports evaluation of the learned navigation strategies. From an ecological perspective, the paths illustrate how agents following a pollution-aware objective move through a landscape of varying AQI; they can be used to formulate hypotheses about how movement might respond to air quality, to be tested against independent movement data. They are simulated paths, not observed bird tracks.

\begin{figure}[htbp]
    \centering
    \includegraphics[width=0.75\textwidth]{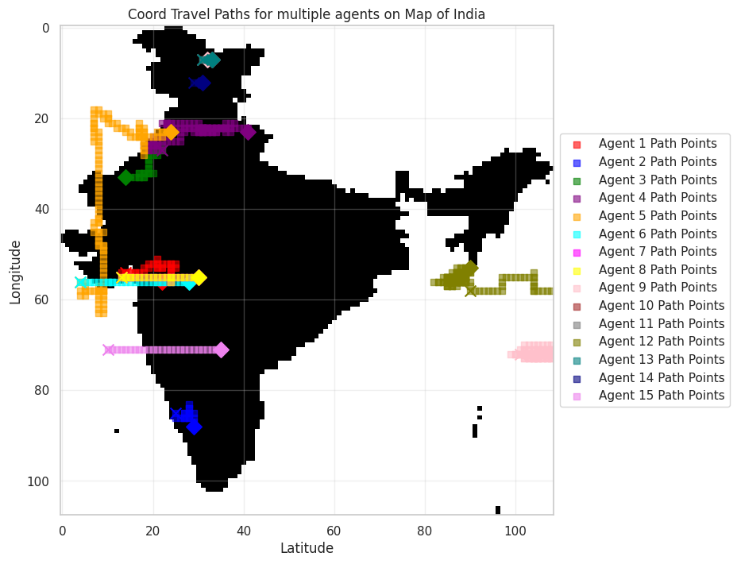}
    \caption{Simulated traversal paths of a subset of 15 agents on the map of India. These are simulated paths under the grid and reward constraints, not observed bird tracks.}
    \label{fig:j16}
\end{figure}

Because more than 400 agents are trained, plotting all their paths in one figure would be unreadable. A subset of 15 agents is therefore shown in Figure~\ref{fig:j16}, each in a distinct colour, with start and end points marked as (x) and (\ding{117}), respectively. The simulation uses an India grid of $108\times109$ cells, each corresponding to a 0.27-degree shift in latitude or longitude, approximately a 30\,km displacement in either direction.

\subsubsection{Analysis of Agent Traversal Paths}

The diversity of agent paths reflects the exploration--exploitation trade-off of reinforcement learning. Initially, agents explore and gather information about regions that may yield higher reward. Over time, behaviour shifts toward exploitation, in which agents favour regions that have produced favourable outcomes, allowing them to maximise cumulative reward while refining their navigation strategies.

This pattern has an ecological analogue: birds explore new environments but tend to return to places where they have found favourable conditions, such as good air quality or abundant food. The agents' shift toward known favourable regions mirrors, at a conceptual level, how birds might adapt movement to environmental cues such as AQI, weather or habitat availability. The interpretations below should be read as hypotheses generated by the simulation rather than as validated ecological findings.

Agents can be grouped by their traversal patterns, which relate to the environmental conditions they encounter:

\begin{itemize}
    \item \textbf{Group A: Coastal Regions}  
    The simulations conducted using the DQN model in coastal regions indicate stable and consistent agent movement patterns. This stability primarily results from relatively constant AQI values, as pollutants in these areas are consistently dispersed by sea breezes. From an ornithological perspective, coastal birds are accustomed to stable environmental conditions, leading to minimal variability in foraging and migration behaviors. This ecological stability is reflected in the simulation, where the agents' trajectories converge toward consistent paths due to predictable resource availability and limited AQI fluctuations.
    
    \item \textbf{Group B: Urban Areas}  
    In urban environments, AQI values fluctuate significantly, leading to highly adaptive and variable agent movement paths. The unpredictability in AQI conditions necessitates real-time decision-making, akin to the complex navigation strategies required in urban settings. From an ecological standpoint, birds in urban habitats must adapt to highly dynamic and often hazardous pollution levels, which fosters behavioral plasticity and erratic movement patterns. This adaptability is mirrored in the DQN model, where agents continuously adjust their navigation strategies in response to the volatile urban landscape to optimize exposure to better air quality.
    
    \item \textbf{Group C: Rural or Forested Regions}  
    In rural and forested regions, consistently low AQI values contribute to stable agent trajectories. The simulations in these areas yield predictable movement patterns, as the minimal pollution levels facilitate more efficient learning of navigation strategies. Similarly, birds inhabiting these regions are conditioned to function in relatively undisturbed environments, resulting in less erratic movement patterns. The congruence between the model’s outcomes and observed bird behavior highlights the strong influence of stable environmental conditions on agent navigation and habitat quality assessments.
\end{itemize}

These groupings enable a comparative analysis of how diverse environmental conditions influence air quality and agent movement patterns. Figure~\ref{fig:j16} illustrates the traversal paths of selected agents, demonstrating the impact of AQI variations on navigation strategies. For instance, Agents 1 and 3 initially follow similar trajectories in northern India before diverging—Agent 1 moves westward into a region with relatively stable AQI, while Agent 3 navigates toward the eastern region, characterized by greater AQI variability. The DQN algorithm learns these variations and formulates distinct navigation strategies for each agent. Similarly, Agents 7 and 14, moving along coastal regions, exhibit stable movement patterns, reinforcing the model’s preference for paths with minimal AQI fluctuations. Conversely, Agents 10 and 15, navigating central India, demonstrate extensive traversal and adaptive movement strategies in response to highly dynamic AQI conditions. These results highlight how the DQN model captures environmental variations and adjusts agent trajectories accordingly. A more intricate analysis of agent traversal paths allows for several critical deductions:

\begin{itemize}
    \item \textbf{Diversity of Agent Paths}  
    The variation in agent paths reflects diverse exploration and exploitation strategies. This diversity parallels migration patterns in bird species such as the European Swallow, where differences in air quality and resource availability drive variations in migration routes. The observed trajectories provide insight into how agents adapt to AQI variations and geographical constraints.
    
    \item \textbf{Revisitation of Locations}  
    Agents revisiting previously explored locations may indicate regions with high AQI variability. This behavior resembles the tendency of birds to return to areas with inconsistent food sources or fluctuating environmental conditions, reinforcing the relevance of the RL model in ecological studies.
    
    \item \textbf{Performance Assessment of the DQN Model}  
    The analysis of agent paths offers a means to evaluate the efficiency of the DQN learning process. Well-optimized trajectories demonstrate strategic navigation with minimal deviations, akin to how migratory birds optimize routes across seasons. Additionally, adaptive responses to AQI fluctuations and geographical boundaries reflect the model’s ability to dynamically adjust, similar to how birds alter migration routes based on environmental changes.
    
    \item \textbf{Environmental Insights from Agent Paths}  
    Correlating agent movement with AQI data provides valuable environmental insights. For instance, avoidance of specific regions may indicate areas of high pollution, akin to how birds circumvent polluted habitats. Habitat quality assessments can be derived by tracking agent interactions with the environment, where frequent penalties or route deviations may signal adverse ecological conditions.
    
    \item \textbf{Identification of Key Locations}  
    The traversal path analysis helps identify critical regions, such as penalty zones where agents consistently experience negative reinforcement, suggesting challenging terrain or poor air quality. Conversely, zones where agents receive sustained positive rewards resemble ideal habitats with optimal living conditions. These findings contribute to targeted environmental monitoring and conservation strategies.
\end{itemize}

Overall, the traversal-path analysis shows how the RL agents interact with the environmental variables of the world-grounded environment. It improves the interpretability of the DQN model and suggests ecological hypotheses about pollution-aware movement that can be examined with independent data.

\subsubsection{Learning Error Analysis}\label{sec:loss}

\begin{figure}[htbp]
    \centering
    \includegraphics[width=0.75\textwidth]{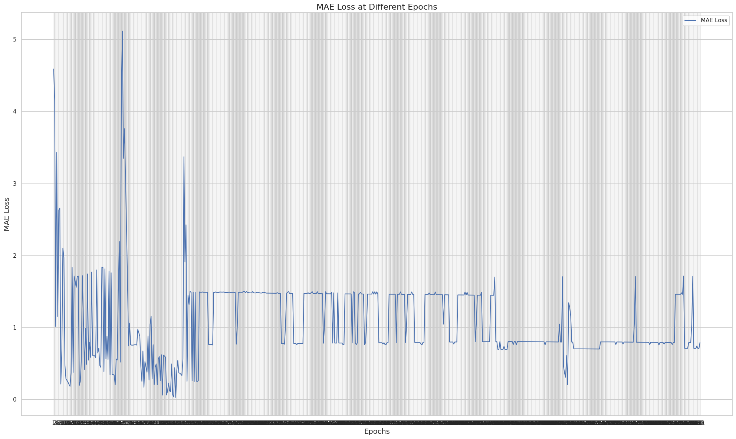}
    \caption{Mean absolute error of the agent model for the first 10 agents across 610 steps; plateaus mark saturation points followed by resets.}
    \label{fig:j17}
\end{figure}

Figure~\ref{fig:j17} presents the plot of the Mean Absolute Error (MAE) of the agent model against the steps taken by each agent, capturing the distribution of approximately 10 agents over a total of 610 steps in the environment. The plot reveals multiple plateau-like curves, indicating points where the agent's learning or adaptation slows down significantly. These plateaus correspond to saturation points, where the agent temporarily ceases to gain new information. For example, when negative reward criteria are applied to regions outside India or near water bodies, agents may continue exploring those areas for a predefined number of steps before resetting to a new location. This reset mechanism, visible in the stopping curves within plateau regions, marks the transition of agents toward exploring different areas. 

The presence of these plateau curves highlights the fundamental exploration-exploitation trade-off inherent in the model. Exploration reflects an agent’s tendency to search for new regions and gather additional information, while exploitation involves leveraging prior knowledge to maximize rewards. The appearance of these plateaus suggests that agents have effectively navigated all known regions, optimizing their learning to achieve peak performance. This balance between exploration and exploitation facilitates efficient adaptation to dynamic environments. 

The reported average MAE is 1.0716, with a minimum MAE of 0.0291 and a maximum MAE of 5.1144. These values quantify the error in the agent model, reflecting some variation in learning performance while maintaining a relatively low average error rate. The MAE plot thus provides critical insights into the learning dynamics of the agents, demonstrating their ability to evolve and adjust to environmental conditions. The observed saturation points and subsequent resets reinforce the model’s capacity to maintain an effective balance between exploration and exploitation while operating in a complex environment.

\subsection{Evaluation Protocol and Validity Considerations}

The comparisons above are descriptive evaluations under the reported simulation setup. Cumulative reward and reward per timestep quantify the specified objective rather than biological fitness; distance measures grid traversal rather than validated flight distance; and smoothness within a run is not evidence of reproducibility across independent runs. The current report does not provide the number of independent training seeds or between-run uncertainty estimates, so the algorithm rankings remain provisional. A stronger evaluation would repeat the full training and evaluation procedure across independent seeds, report seed-level means with 95\% confidence intervals, and match initial-location distributions, rewards, transition rules, episode horizons and evaluation budgets across methods. Because Value Iteration uses an explicit transition model, its information access and convergence tolerance should also be documented. Useful sensitivity checks include the reward coefficients, repetition penalties, agent removal, grid resolution and AQI imputation, and a direct comparison of the mean AQI encountered by each method would help distinguish pollution avoidance from optimisation of the reward-shaping terms. These checks are requirements for further validation rather than experiments claimed here.

\section{Comparative Analysis with Related Works}\label{sec:related}

This section presents a qualitative comparative analysis of existing frameworks reviewed in the literature. Due to the novelty of the proposed approach and the lack of a standardized quantitative evaluation framework, the comparison is conducted based on three key aspects: bioindicator studies, computational simulation, and reinforcement learning (RL).

\subsection{Bioindicator Studies}

Table~\ref{tab:bioindicator_comparison} provides a comparative overview of various bioindicator studies relevant to the proposed work, evaluating their contexts, specific indicators, data collection methodologies, and inherent challenges.

\begin{table}[htbp]
    \centering
    \small
    \caption{Comparison with bioindicator studies.}
    \label{tab:bioindicator_comparison}
        \begin{tabular}{|p{2.2cm}|p{2.9cm}|p{3.0cm}|p{3.4cm}|p{3.4cm}|}
        \hline
        \textbf{Study} & \textbf{Context} & \textbf{Specified Indicators} & \textbf{Data Collection Method} & \textbf{Challenges} \\ \hline
        Singh et al. & Environmental Monitoring & Data Openness, ML Algorithms & Aggregated multi-source data & Data availability, quality issues \\ \hline
        Zhang et al. & Ecosystem Monitoring & Drone Relay, IoRT & Drones, IoRT devices & Connectivity issues, high energy use \\ \hline
        Iverson et al. & Migration Ecology & Land Cover, NDVI, GPS Tracking & GPS tracking, NDVI satellite data & Small bird tracking, data accuracy \\ \hline
        Cunha Silva et al. & Disease Ecology & Habitat Selection, Rabies Risk & Field observations, GPS, surveys & Dog behavior variability, rural data gaps \\ \hline
        Hill et al. & Urban Ecology & Urbanization, Roost Sites & GPS tracking, field observations & Urban accessibility, observer bias \\ \hline
        \textbf{Proposed Work} & Environmental Health Monitoring & AQI, Geographic Data, Bird Calls & API-based bird data, measured AQI, geolocation & Computational cost, static AQI snapshot \\ \hline
    \end{tabular}
\end{table}

A distinguishing feature of the proposed framework is the use of reinforcement learning, specifically Deep Q-Networks, to model movement as an adaptive, interaction-driven process in a world-grounded environment. Unlike Singh et al., who used machine learning for data analysis without an interactive model, the proposed approach embeds measured data in an environment in which agents learn from feedback. While Singh et al.'s method benefited from data openness, it was constrained by data availability and quality; the proposed framework mitigates the cost of repeated data access through its precomputed AQI layer.

Zhang et al. used drones and the Internet of Remote Things (IoRT) for continuous environmental monitoring. Despite high-resolution sensing, this approach is constrained by energy consumption and connectivity in remote areas. By contrast, the proposed study relies on AQI data and bird-recording metadata that are accessible through existing APIs and require no field infrastructure, and its reinforcement learning model adds a simulation capability that sensing alone does not provide.

Studies such as Iverson et al. and Cunha Silva et al. used GPS tracking and field observation to analyse habitat selection and migration. These methods are effective for localised studies but face scalability and accuracy limits in remote or inaccessible regions, and they do not provide a simulation model for exploring scenarios. The proposed multi-instance RL framework integrates AQI, geospatial parameters and bird-recording metadata to simulate pollution-aware movement at country scale, complementing such observational data, which remain essential for validation.

Hill et al. studied vulture responses to urbanisation with spatial analysis, facing accessibility constraints in urban environments and potential observer bias. The proposed work instead combines AQI data, geolocation and reinforcement learning to simulate movement. While computationally intensive, this approach offers a scalable complement to field methods for environmental health monitoring.

By addressing limitations in data collection, scalability and modelling found in prior bioindicator studies, the proposed work contributes a data-driven simulation approach that can support ecological monitoring and environmental health assessment.

\subsection{Computational Simulation Methods}

Table~\ref{tab:simulation_comparison} evaluates different studies based on the simulation methodologies they employed, with a focus on the metrics used, results obtained, and the challenges encountered.

\begin{table}[htbp]
    \centering
    \small
    \caption{Comparison with computational simulation studies.}
    \label{tab:simulation_comparison}
    \begin{tabular}{|p{2.2cm}|p{3.0cm}|p{3.0cm}|p{3.4cm}|p{3.3cm}|}
        \hline
        \textbf{Paper} & \textbf{Proposed Methodology} & \textbf{Metrics Used} & \textbf{Results} & \textbf{Challenges} \\ \hline
        Saulnier et al. & Statistical Analysis, Climate Models & Reproductive Success Rates, Environmental Correlations & Significant differences in reproductive success & Data variability, modeling diverse conditions \\ \hline
        Antelmi et al. & Agent-Based Models (ABM) & Simulation Accuracy, Agent Behavior Patterns & Captures complex interactions at a micro level & Computational intensity, scalability issues \\ \hline
        Perry et al. & Deep Learning, Neural Networks & Model Accuracy, Predictive Power & Improved ecosystem modeling & High data requirements, model interpretability \\ \hline
        Tovey et al. & Multi-Agent Reinforcement Learning (MARL) & Strategy Effectiveness, Adaptability & Models complex adaptive behaviors & High training data demands, computational complexity \\ \hline
        Tappler et al. & Continuous Stochastic Dynamics, RL & Predictive Accuracy, Adaptability & Captures dynamic environmental processes & High computational cost, complex implementation \\ \hline
        Ellis et al. & Data Integration, Machine Learning & Habitat Use Patterns, Predictive Accuracy & Insights into habitat use patterns & Data integration complexity, data variability \\ \hline
        \textbf{Proposed Work} & World-grounded multi-instance RL, DQN, AQI and bird-recording integration & Cumulative reward, diversity, step efficiency, distance, MAE & Highest reward and step efficiency among compared RL methods (single runs) & Computational burden, static AQI, no trajectory validation \\ \hline
    \end{tabular}
\end{table}

Previous research has used computational methods that face challenges in scalability and complexity. For instance, Antelmi et al. applied agent-based modelling (ABM) to simulate micro-level interactions within ecosystems, but ABM faced scalability issues that limit its use for large systems. Our study, though computationally intensive because of its RL setup, reduces data-collection cost by using existing APIs---Xeno-canto for bird-recording data and Ozon3 for air quality---instead of extensive field collection, and reduces runtime cost through the precomputed AQI matrix.

Several studies, including those by Perry et al. and Ellis et al., applied deep learning and statistical models to ecological systems. These methods improved model accuracy but were hindered by complex data interpretation, high computational cost and difficult data integration. Perry et al. highlighted large data requirements and the complexity of neural network outputs as key obstacles, and Ellis et al. integrated diverse data sources to study habitat use, which made the approach cumbersome. Our work keeps data integration simple by using well-structured, readily available data such as AQI and bird-recording metadata, and the DQN within a multi-instance framework exposes reward-based metrics that are straightforward to interpret.

Many studies that used machine learning or statistical models were limited in adaptability. Tappler et al. used RL with continuous stochastic dynamics to simulate environmental processes; the method was adaptable but required high computational power at scale. Saulnier et al. applied statistical analysis to relate environmental factors to bird reproductive success; their results highlighted important influences but did not provide an adaptive model. In contrast, the proposed work models movement as an interactive learning process, in which the DQN adapts to the environmental feedback it receives and the system can simulate different environmental scenarios. With time-resolved AQI inputs, the same structure could be updated as conditions change.

In ecological contexts, studies such as those of Iverson et al. and Cunha Silva et al. faced scalability and applicability issues when extended to broader settings. Our framework integrates environmental and ecological data in one model and is designed to accept data from different regions and geographic scales, including air-quality indices, geographic information and species metadata, so it can be applied across urban and rural settings and, with suitable data, to other species and processes.

\subsection{Reinforcement Learning-Based Modelling}

Table~\ref{tab:rl_comparison} evaluates various studies based on the reinforcement learning (RL) methodologies employed, focusing on their specific RL approaches, key metrics, results, inferences, and limitations.

\begin{table}[htbp]
    \centering
    \small
    \caption{Comparison with reinforcement learning-based studies.}
    \label{tab:rl_comparison}
    \begin{tabular}{|p{1.9cm}|p{2.5cm}|p{2.3cm}|p{3.0cm}|p{2.8cm}|p{2.6cm}|}
        \hline
        \textbf{Paper} & \textbf{Proposed RL Methodology} & \textbf{Metrics Used} & \textbf{Results} & \textbf{Inferences} & \textbf{Cons} \\ \hline
        Haensel et al. & Agent-Based Modeling (ABM) & Student Outcomes in Complexity & ABM aids teaching of human-environment interactions & Suitable for complex system simulations in education & Simplistic model, limited real-time adaptability \\ \hline
        Martínez et al. & DRL for Autonomous Flocking & Convergence, Stability, Flocking Behavior & DRL optimizes coordination and navigation in flocking & Adaptable to dynamic environments, real-time effectiveness & Computationally expensive, lengthy training, complex tuning \\ \hline
        Tovey et al. & Micro-scale Multi-Agent RL (MARL) & Strategy Emergence, Adaptability & MARL models adaptive behavior under diverse conditions & Suitable for emergent behavior modeling in dynamic systems & High training data needs, high computational cost \\ \hline
        Tappler et al. & Continuous Stochastic Dynamics in RL & Model Accuracy, Predictive Performance & Captures dynamic environmental processes effectively & Supports continuous learning in environmental systems & High computational cost, complex implementation \\ \hline
        Li et al. & Ensemble RL Gated Unit Model & PM2.5 Forecasting Accuracy & Improved air quality forecasting with ensemble learning & Effective for environmental predictions & Complex model, high data requirements \\ \hline
        \textbf{Proposed Work} & Multi-instance DQN in a world-grounded AQI environment & Reward, diversity, efficiency, distance, MAE & Peak per-step reward of 722; highest cumulative reward among compared methods & Pollution-aware movement simulation grounded in measured AQI & High computational needs; static AQI; validation pending \\ \hline
    \end{tabular}
\end{table}

The proposed study embeds Deep Q-Networks in a multi-instance reinforcement learning framework designed to simulate bird propagation driven by environmental variables, in particular the Air Quality Index. This extends the agent-based modelling (ABM) of Haensel et al., which was effective for teaching human--environment systems but did not integrate measured environmental data, which is essential for world-grounded simulation. Haensel et al.'s ABM focused on representing interactions within land-use policies rather than environmental interactions at scale, and the simplifications of rule-based ABM restrict its use for modelling the environmental variation that influences movement. In the DQN-based approach, agents learn navigation policies from measured AQI data rather than from hand-specified rules. The use of bird-recording metadata from Xeno-canto further anchors the simulation in real-world observations.

A key feature of this work is the integration of diverse datasets---geographic coordinates, AQI values and bird-recording metadata---into a single multi-instance simulation. Martínez et al., who used Deep Reinforcement Learning (DRL) for autonomous robot flocking, demonstrated the suitability of DRL for dynamic decision-making, achieving good stability and adaptability at the cost of high computational demand and extensive parameter tuning. The proposed DQN-based method shares the challenge of computational intensity, while addressing a different, ecologically motivated problem in which the effect of air quality on movement is explicitly modelled.

Whereas Martínez et al. focused on coordination among robotic agents, our multi-instance DQN framework trains many agents, each initialised from a different recording location, as they respond to the AQI landscape across a large geographic region. This makes it possible to study how regional variation in air quality shapes simulated movement. Li et al. used an ensemble RL method for daily PM2.5 forecasting; although their work advanced air-quality prediction, it did not model the effect of air quality on animal movement, which is the focus here.

A general limitation of RL-based methods is their computational cost, which the proposed model shares. Martínez et al., Tovey et al. and Tappler et al. all highlight the cost of RL techniques on large datasets or environments, and our multi-instance framework likewise requires substantial resources to simulate many agents across a large region.

To mitigate these costs, the framework uses several optimisations. AQI data are precomputed and stored on a grid, so that no API calls are needed during simulation, and the geographic space is restricted by masking to valid cells of the study region. This improves scalability without changing the environment's structure. Compared with studies such as Li et al. and Tappler et al., the framework aims for a practical balance between model complexity and scalability, particularly in data retrieval.

\section{Discussion}\label{sec:discussion}

AI-driven simulations, in particular those based on reinforcement learning, offer a structured and scalable approach to modelling ecological dynamics that complements traditional observational methods. Concerns that such models omit real-world ecological factors are legitimate; the world-grounded design addresses part of this by building the environment from domain data---AQI, bird-recording metadata and geographic structure---while the remaining factors are discussed in Section~\ref{sec:limitations}. Unlike approaches based on static observations and small samples, the framework allows large-scale, data-driven exploration of possible interactions between movement and air quality. Reinforcement learning does not replace ecological expertise; it complements it by allowing hypotheses to be explored in a controlled and reproducible setting, and the framework can be refined as new ecological data are incorporated.

The framework has potential implications for urban planning and pollution control by providing a data-driven way to explore the ecological relevance of air-quality variation. Simulating movement in response to AQI can help identify candidate pollution hotspots and regions whose habitat value may be at risk, to be examined with field data. Such analyses could inform targeted mitigation, such as green-space planning, buffer zones and emission controls in areas important for avian biodiversity, and could feed into urban planning that aims to minimise ecological disruption.

The framework uses a holistic multi-instance reinforcement learning design to simulate movement across India. Although the implementation overlays an AQI mask on India, the framework is adaptable to other regions. The AQI layer shapes where agents can profitably move, and the DQN selects actions by predicting Q-values to maximise cumulative reward. The framework visualises traversal paths and cumulative rewards, showing how the DQN agents adjust their paths to AQI and geographic constraints while balancing exploration and exploitation. The learning-error plot reflects the adaptive nature of the setup, with plateaus indicating exploration of new areas alongside reinforcement of learned behaviour, and the average MAE of 1.0716 over the first ten agents summarises the stability of the agent model during learning. Together these results show that DQN can learn effective policies for the specified pollution-aware objective in a world-grounded environment.

The framework addresses several limitations identified in the literature. Satellite imagery and on-ground sensors face high costs, limited coverage and delays. GPS tags on species such as the Golden-crowned Sparrow provide valuable movement data but are expensive and difficult to deploy at scale. Remote-camera studies of vulture roost sites give insight into urbanisation's effect on behaviour but do not model environmental interactions. Inspired by data-acquisition systems such as drone-enabled IoRT, the framework integrates diverse data sources into one simulation. Its DQN addresses the limitations of supervised and unsupervised methods identified in Section~\ref{sec:background} by modelling movement as adaptive behaviour in response to AQI and geographic constraints.

The methodology can also be extended to other environmental questions. Avian movement as a bioindicator could be adapted to aquatic ecosystems by simulating waterfowl responses to water-quality variation, including pollution and vegetation cover, which would be useful in wetlands and coastal environments where conventional monitoring is logistically difficult. Multi-instance simulations with measured data could also support urban wildlife management by exploring the effects of green-space availability, noise and air quality on bird populations, informing ecologically sensitive infrastructure. In agricultural landscapes, simulated animal movement could help with early detection of pest pressure and crop health issues. These extensions illustrate the versatility of world-grounded reinforcement learning as a basis for data-driven environmental management.

\section{Limitations, Validation and Future Work}\label{sec:limitations}

\subsection{Limitations}

While the framework offers a new approach, it has several data and methodological limitations. First, the AQI dataset is static, representing a snapshot in time rather than a dynamically updated series; its results therefore concern navigation within a fixed AQI landscape, not responses to temporal environmental change. Dynamic AQI data would improve fidelity but increase computational demands. Second, the Xeno-canto records are occurrences rather than tracks, so the generated paths cannot be validated against observed trajectories. Third, the current evaluation does not report independent training seeds or between-run uncertainty, so algorithm rankings remain provisional. Fourth, the reward expressions in Section~\ref{sec:method} are not equivalent, and the implemented reward, its coefficients and the timing of AQI evaluation should be fixed for each method. Fifth, agent removal changes the evaluated population.

For visual clarity, path projections are shown for a subset of agents only, and learning-error curves for the first ten agents, since plotting the full set would be difficult to interpret; this selective presentation helped identify saturation points and the loss trends. The overall reinforcement learning process was computationally intensive and required long execution times. Although AQI is an important environmental factor, relying on it alone is limiting: bird movement and habitat selection depend on many variables, including food availability, climate, predation risk and habitat structure, which AQI does not capture. AQI is also a generalised metric that may not reflect microhabitat-level air quality, which limits precision for fine-scale responses, and seasonal variation and sudden pollution spikes may cause transient behavioural shifts that are difficult to generalise. Despite these limitations, AQI is a useful environmental signal, and combined with additional ecological parameters it can support a computationally efficient framework for studying pollution-driven movement.

In the model, agents are treated as independent entities per record, which may overlook species-specific behaviour. Involving ornithologists and incorporating richer behavioural data would help refine the simulations. Expanding geographic coverage, integrating time-resolved data and using more advanced algorithms could further improve the model, and other bioindicator species such as butterflies and bees could give a more comprehensive view of ecosystem health. The method remains adaptable and serves as a starting point for multidisciplinary investigation.

\subsection{Validation Requirements}

Before ecological conclusions can be drawn, the following validation steps are required: (i) repeating the complete training and evaluation across independent seeds and reporting seed-level means with 95\% confidence intervals; (ii) matching initial-location distributions, rewards, transition rules, horizons and evaluation budgets across methods; (iii) sensitivity analyses over reward coefficients, repetition penalties, agent removal, grid resolution and AQI imputation; (iv) reporting the mean AQI encountered by each method to separate pollution avoidance from reward shaping; and (v) testing against timestamp-aligned AQI sequences and independent movement-tracking data with species-specific habitat and seasonal information.

\subsection{Future Work}

Future work should focus on further refining our simulation model to develop more holistic and realistic representations of bird propagation and environmental monitoring. The simulation environment could incorporate additional environmental variables, such as temperature, humidity, vegetation cover, water availability, and noise pollution, to enhance the realism of modeled bird behaviors and align them more closely with real-world scenarios \cite{Colin2023,Sihao2023}. While the current demonstration used India as the test case, the framework is easily adaptable to other countries, and expanding it to different regions would further test its strength and versatility. This would help better understand migratory behaviors and habitat preferences across various ecological zones \cite{Migration2022,Huang2022}. For instance, applying the model in regions like North America or Europe could showcase how birds adapt to differing climatic conditions, land use patterns, and further our understanding of global migratory patterns. Integrating time-resolved data, such as live AQI feeds, would improve the model’s relevance and accuracy. Real-time data integration would allow for immediate feedback on how changes in pollution and other environmental factors influence bird movements, offering an opportunity to enhance adaptive conservation strategies \cite{Mutalemwa2023,Timm2023,Keren2023}. This could be achieved by developing real-time environmental monitoring stations that directly feed data into the model, ensuring that predictions and analyses remain current and actionable. Furthermore, it would be essential to model how different species respond to changes in their environments, providing valuable insights into ecosystem health and biodiversity \cite{Laender2023,Rigal2023}. This would enable simulations of how various bird species interact with their environments in an integrated manner. The results from such scenario analyses, along with input from stakeholder consultations, could guide the development of policy and conservation strategies that are practical and culturally relevant \cite{Gianicolo2021,Monniez2022}.  The simulation framework could also be used to test selected conservation scenarios and policies, working collaboratively with local communities, governments, and conservation organizations to implement the most effective strategies. Enhancing visualization and creating interactive user interfaces, such as augmented reality (AR), could help illustrate bird movement patterns and environmental changes more vividly \cite{Waghmare2016,Veas2013,Kim2018}. AR could provide interactive maps and simulations, making the impact of bird migration and environmental factors more engaging and educational. This approach would help increase public participation in conservation efforts by making such information more accessible and relevant to local communities.

\section{Conclusion}\label{sec:conclusion}

This paper presented a world-grounded, multi-instance reinforcement learning framework for pollution-aware bird movement modelling within a geographically constrained grid of India. A precomputed AQI matrix, built from sampled air-quality data with neighbour imputation and aligned with a rasterised state map, supplies measured environmental inputs to the agents' reward, and more than 400 DQN agents initialised from Xeno-canto recording locations learn to navigate this environment. In comparisons with Monte Carlo, Q-Learning and Value Iteration, DQN attained the highest cumulative reward and step efficiency in the reported runs, with a peak per-step reward of 722 and an average MAE of 1.0716 over the first ten agents. Traversal-path and learning-error analyses illustrate how the agents balance exploration and exploitation under environmental constraints.

The Xeno-canto occurrences provide geographic grounding but do not validate the generated movement paths, and the present evaluation does not establish robustness across independent training seeds. The contribution is therefore a computational proof of concept and a decision-support basis for environmental monitoring. Explicit reconciliation of the implemented reward, uncertainty-aware evaluation, and validation against time-resolved environmental and movement data are the next steps toward conclusions about avian behaviour or ecological outcomes. Extending the environment with additional variables and regions would further broaden its use in conservation planning, environmental policy and ecological research.

\section*{Note on This Version}
A condensed version of this work, titled \emph{World-Grounded Reinforcement Learning for Pollution-Aware Bird Movement Modeling Using a Multi-Instance DQN Framework}, has been accepted at the GlobalSouthAI Affinity Event at NeurIPS 2026 (\url{\venueurl}). Because of the workshop page limit, that paper presents the framework and main comparative results in condensed form. This arXiv preprint is the extended version: it gives the complete literature review and problem formulation, the full AQI-mapping, environment, DQN and runtime algorithms, the data description and exploratory analysis, the detailed per-metric comparison and traversal-path analysis, the learning-error analysis, the qualitative comparison with related work, and an expanded discussion of limitations and validation. This preprint has not been peer reviewed in this extended form.

\section*{Availability of Data and Materials}
The external datasets and tools used in this study are referenced in the manuscript: AQI mapping was performed with the Ozon3 library (\url{https://github.com/Ozon3Org/Ozon3}), state-level shapefiles were obtained from \url{https://www.data.gov.in}, and bird-recording data were retrieved from \url{https://xeno-canto.org}. Procured and processed data are available from the authors on reasonable request.

{\small

\end{document}